\documentclass[letterpaper,journal]{IEEEtran}

\usepackage{amsmath,amsfonts,amssymb}
\usepackage{algorithm}
\usepackage{algorithmic}
\usepackage{array}
\usepackage[caption=false,font=normalsize,labelfont=sf,textfont=sf]{subfig}
\usepackage{textcomp}
\usepackage{stfloats}
\usepackage{url}
\usepackage{verbatim}
\usepackage{graphicx}
\usepackage{cite}
\usepackage{booktabs}
\usepackage{multirow}
\usepackage{adjustbox}
\usepackage{xcolor}
\usepackage{colortbl}
\usepackage{afterpage}
\usepackage{pdfpages}

\definecolor{baselineblue}{RGB}{0,82,155}
\definecolor{darkred}{RGB}{178,34,34}

\usepackage{enumitem}

\begin{document}

\title{$\mathcal{PC}^{2}$-AD: Point Cloud Upsampling to Safeguard 3D Anomaly Detection with Resolution-constrained Edge Devices}

\author{
Yutong Gu, Yingxi Xie, Kejin Huang, Jian Ning, Hanzhe Liang$^{\dagger}$,
\IEEEmembership{Member, IEEE},\\
Linlin Shen, \IEEEmembership{Senior Member, IEEE},
Jinbao Wang, \IEEEmembership{Member, IEEE}

\thanks{This work was supported in part by 
the Signal, Information, and Biological System Processing Laboratory at Shenzhen Audencia Financial Technology Institute, Shenzhen University, the National Natural Science Foundation of China (Grant No. 62576218), Guangdong Provincial Key Laboratory (Grant No. 2023B1212060076), the Intelligent Computing Center of Shenzhen University (Corresponding author: Hanzhe Liang)}

\thanks{Yutong Gu, Kejin Huang, and Yingxi Xie are with the Shenzhen University WeBank Institute of Finance, Shenzhen University,
Shenzhen 518060, China (e-mail: 2025290265@mails.szu.edu.cn; 2025290244@mails.szu.edu.cn; 2025290209@mails.szu.edu.cn).}

\thanks{Jian Ning is with the School of Computer Science, Wuhan University, Wuhan 430072, China
(e-mail: ningjian@whu.edu.cn).}

\thanks{Hanzhe Liang is with the Shenzhen Audencia Financial Technology Institute, Shenzhen University, Shenzhen 518060, China, and also with the Mohamed bin Zayed University of Artificial Intelligence, Masdar City, Abu Dhabi, UAE (e-mail: Hanzhe.Liang@mbzuai.ac.ae).}

\thanks{Jinbao Wang and Linlin Shen are with the School of Artificial
Intelligence, Shenzhen University, Shenzhen, Guangdong, China
(e-mail: wangjb@szu.edu.cn; llshen@szu.edu.cn).}

}

\maketitle

\begin{abstract}

Low-cost and low-resolution sensors used in edge deployments can produce test point clouds that are substantially sparser than the normal training data. This train--test sampling-resolution gap changes the local geometry available to a 3D anomaly detector. We propose $\mathcal{PC}^{2}$-AD, a point cloud upsampling framework that compensates sparse test inputs before downstream detection. Target Domain Candidate Generation (TCG) adapts a pretrained upsampler to normal training geometry and generates a dense candidate pool. Geometry-Aware Candidate Filtering (GACF) selects candidates according to geometric spacing and spatial coverage. Normality-Preserving Point Compensation (NPPC) refines the selection by comparing candidate normality scores with those of their input anchors. The selected points are combined with the unchanged input points and processed by the existing detector. Experiments with six detectors on two Anomaly-ShapeNet settings and Real3D-AD show improvements in the mean of object-level and point-level AUROC for all six detectors in each Anomaly-ShapeNet setting and four on Real3D-AD. These results support point cloud compensation as an input-level approach to improving 3D anomaly detection under low-resolution sensing conditions. Code is publicly available at \url{https://github.com/gyutong406-commits/PC2-AD}.

\end{abstract}


\section{Introduction}

\begin{figure}[t]
    \centering
    \includegraphics[width=\linewidth]
    {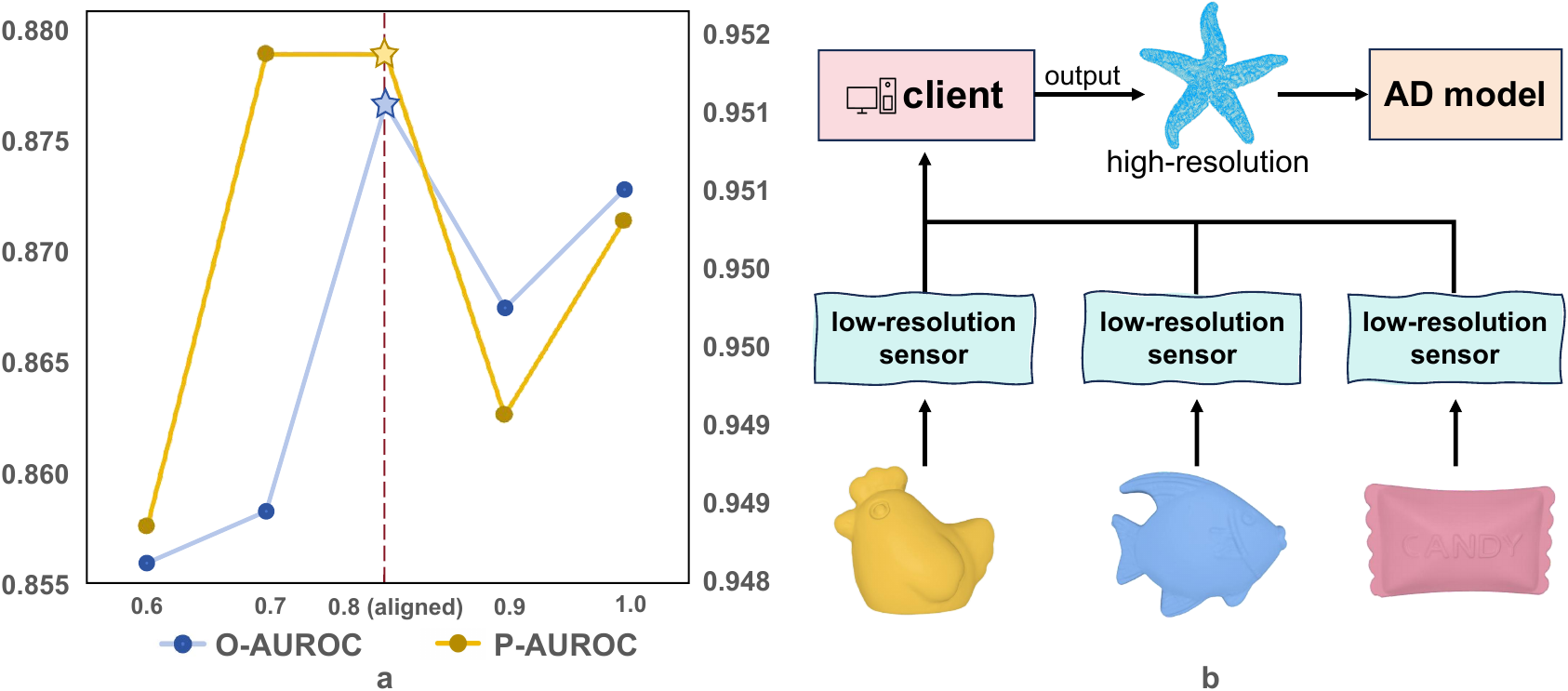}
    \caption{Motivation for client-side resolution compensation. (a) O-AUROC and P-AUROC under different test resolutions, with the training resolution fixed at $0.8$. The dashed line indicates the aligned setting. (b) Low-resolution sensor inputs are compensated on the client before being passed to the 3D anomaly detection model.}
    \label{fig:motivation}
\end{figure}

Three-dimensional anomaly detection~(3D-AD) identifies anomalies and localizes anomalous regions in 3D data represented as point clouds~\cite{11644920,yu2025reg2inv}, depth maps~\cite{3746027.3755261,Long_2026_CVPR}, or meshes~\cite{11445823}. 
Owing to the ability of 3D representations to mitigate occlusion and capture detailed geometric information, 3D anomaly detection holds significant practical value in industrial quality inspection~\cite{liu2023neurips-real3dad,cheng2025mc3dad}.
Existing 3D anomaly detection methods predominantly adopt self-supervised learning, a tendency mainly attributed to the reliance on high-resolution sensors and the substantial time required to acquire high-precision point clouds~\cite{liu2023neurips-real3dad}. These methods can be broadly categorized into feature-embedding methods, feature-reconstruction methods, and other approaches.

Feature-embedding methods employ a pretrained feature extractor to represent normal training samples and construct normal prototypes. The same extractor is applied to test samples, and anomalies are detected based on their feature discrepancies from the normal prototypes~\cite{roth2022patchcore,yu2025reg2inv}. Reg3D-AD~\cite{liu2023neurips-real3dad} uses RANSAC~\cite{1623272} for point cloud registration and combines Point-MAE features~\cite{pang2022masked} with point coordinates to obtain comparable representations, while Reg2Inv~\cite{yu2025reg2inv} extracts rotation-invariant features to obtain robust representations for 3D anomaly detection. Meanwhile, CPMF~\cite{cao2024complementary} and ISMP~\cite{liang2025look} leverage projected pseudo-modalities to capture additional complementary information.
These methods rely on feature comparability, requiring the training and test samples to share a consistent feature space~\cite{GLFM}. Moreover, normal and anomalous features need to be decoupled at both the geometric and semantic levels~\cite{ijcai2025p182,roth2022patchcore}.

Feature-reconstruction methods learn a mapping that accurately reconstructs normal features while producing larger reconstruction errors for anomalous features~\cite{cheng2025mc3dad,liang2025taming}. MC3D-AD~\cite{cheng2025mc3dad} and IRM3D-AD~\cite{11644920} train an encoder–decoder architecture and detect anomalies based on feature-level reconstruction errors. PO3AD~\cite{PO3AD} and CASL~\cite{zha2026casl} directly predict point-wise offsets to detect anomalies and restore anomalous regions.
These methods typically rely on anomaly synthesis or carefully designed noise injection schemes, which must be tailored to different data sources~\cite{balapour2026anomalyfactory3dmodular}.

The above 3D anomaly detection methods are sensitive to point cloud resolution, and their perception of anomalies varies across scales~\cite{cheng2026towards,664647.3680919}. This typically requires high-resolution input point clouds drawn from the same domain as the training data~\cite{GLFM}. These requirements are overly stringent, particularly for practical edge deployment, where high-resolution point-cloud sensors are costly, and data acquisition is time-consuming. To validate this observation, we conduct controlled experiments on a subset of Anomaly-ShapeNet. As shown in Fig.~\ref{fig:motivation}(a), the aligned test resolution achieves the highest O-AUROC among the evaluated settings, suggesting that reducing the train--test resolution gap can benefit 3D anomaly detection.
\textbf{Therefore, low-cost and low-resolution sensors used in edge deployments may degrade detection performance, whereas high-resolution sensors incur greater hardware costs and processing overhead.} Recovering faithful high-resolution point clouds from low-resolution measurements while facilitating downstream anomaly detection would therefore greatly advance the practical deployment of 3D anomaly detection.

To this end, we propose $\mathcal{PC}^{2}$-AD, whose overall workflow is illustrated in Fig.~\ref{fig:motivation}(b). We first use Target Domain Candidate Generation (TCG) to adapt a pretrained point cloud upsampling model to the normal geometry of the target domain and generate a dense candidate pool for each low-resolution test point cloud. Geometry-Aware Candidate Filtering (GACF) then selects geometrically reliable and spatially well-distributed candidates, while Normality-Preserving Point Compensation (NPPC) further refines the selected candidates according to their local normality consistency with the original observations. The resulting compensation points are finally combined with the original test point cloud and passed to a downstream 3D anomaly detector. 
The main contributions are summarized as follows:

\begin{itemize}
    \item To the best of our knowledge, we are the first to identify and systematically investigate the train--test sampling-resolution gap in 3D anomaly detection, which arises when models trained on high-resolution normal point clouds are deployed with low-cost, low-resolution sensors.

    \item We propose $\mathcal{PC}^{2}$-AD, a normality-preserving point cloud resolution compensation framework that uses only normal training data to compensate low-resolution test inputs without modifying the downstream anomaly detector.

    \item We develop target-domain candidate generation, geometry-aware candidate filtering, and normality-preserving point compensation to generate reliable and spatially distributed compensation points while retaining the local anomaly evidence carried by the original observations.

    \item Extensive experiments demonstrate that $\mathcal{PC}^{2}$-AD improves the average of O-AUROC and P-AUROC in most evaluated configurations under low-resolution inference and is compatible with different downstream 3D anomaly detection methods.
\end{itemize}

The remainder of this paper is organized as follows. Section~II reviews related work. Section~III presents $\mathcal{PC}^{2}$-AD. Section~IV reports the experimental results, and Section~V concludes the paper.

\section{Related Work}

\subsection{3D Anomaly Detection}

3D anomaly detection aims to identify anomalous objects and localize abnormal regions in point clouds. Existing methods are primarily categorized into feature embedding and reconstruction or normal structure modeling. Embedding-based methods, including PatchCore~\cite{roth2022patchcore}, Reg3D-AD~\cite{liu2023neurips-real3dad}, Reg2Inv~\cite{yu2025reg2inv}, Group3AD~\cite{664647.3680919}, CPMF~\cite{cao2024complementary}, and ISMP~\cite{liang2025look}, construct normal feature references or memory banks, often leveraging pretrained encoders such as Point-BERT~\cite{yu2022pointbert} and Point-MAE~\cite{pang2022masked}, and detect anomalies through feature discrepancies. Reconstruction-based methods, such as IMRNet~\cite{anomaly_shapenet}, R3D-AD~\cite{zhou2024r3dad}, PO3AD~\cite{PO3AD}, MC3D-AD~\cite{cheng2025mc3dad}, Template3D-AD~\cite{ijcai2025p182}, and SplatPose~\cite{Kruse_2024_CVPR}, model normal geometry through prediction, denoising, template learning, or explicit 3D reconstruction. Some multimodal methods, including M3DM~\cite{wang2023cvpr-multimodal}, 2M3DF~\cite{asad2025_2m3df}, and GPAD~\cite{li_gpad}, also combine appearance and geometric information to enhance anomaly detection. Our study concerns the test point set supplied to a detector: we compensate sparse inputs while keeping the downstream training and model parameters fixed.

\subsection{Point Cloud Upsampling}

Point cloud upsampling aims to increase the sampling density of sparse observations while preserving their geometric structure. PU-Net~\cite{Yu_2018_CVPR}, PU-GCN~\cite{Qian_2021_CVPR}, and PU-Transformer~\cite{Qiu_2022_ACCV} generate additional points through feature extraction and expansion. GC-PCU~\cite{ding2021_gcpcu} learns perturbations to estimate point coordinate shifts. PU-Mask~\cite{liu2024_pumask} uses implicit virtual masks for local filling, while PU-GSM~\cite{liu2025_pugsm} exploits geometry-guided global self-similarity. Grad-PU~\cite{He_2023_CVPR} learns a distance field and iteratively refines point coordinates, supporting flexible upsampling ratios. This capability can compensate sparse sensor measurements when normal training clouds have higher sampling densities. For anomaly detection, the added points also change the local neighborhoods used to distinguish normal and anomalous geometry, making candidate selection part of the resolution compensation problem. Building on Grad-PU, $\mathcal{PC}^{2}$-AD adapts candidate generation to normal training geometry and selects compensation points according to geometric reliability and local normality consistency while retaining the original observations.

\section{Method}

\subsection{Problem Formulation}
Let $\mathcal{P}^{\mathrm{train}}=\{\mathbf{P}^{\mathrm{train}}_i\}_{i=1}^{N}$ be the normal training set from $K$ object categories, where $\mathbf{P}^{\mathrm{train}}_i\in\mathbb{R}^{n_i^{\mathrm{train}}\times3}$. The test set is $\mathcal{P}^{\mathrm{test}}=\{\mathbf{P}^{\mathrm{test}}_i\}_{i=1}^{m}$, with $\mathbf{P}^{\mathrm{test}}_i\in\mathbb{R}^{n_i^{\mathrm{test}}\times3}$. Low-resolution sampling can affect anomaly visibility~\cite{cheng2026towards}. Motivated by the sampling-resolution gap between high-resolution normal training clouds and sparse sensor measurements, we learn a resolution compensation model $\mathcal{R}$ from normal training data. For each test cloud, the model selects additional points and combines them with the unchanged input points before downstream detection. The target point count is specified by the dataset protocol. Category $c$ indexes the normal feature center and feature bank introduced below.

\subsection{Overview}
\begin{figure*}[htbp]
    \centering
    \includegraphics[width=\textwidth]{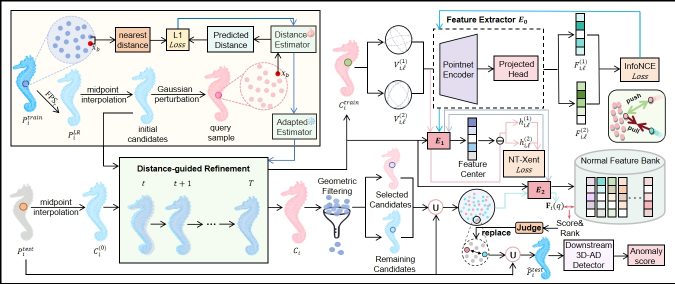}
    \caption{Overview of $\mathcal{PC}^{2}$-AD. TCG adapts a pretrained distance estimator through TDA and uses it for candidate generation (CG). GACF filters the candidates by anchor distance and spatial coverage. NPPC trains the patch encoder through CNP and MSCF, constructs a normal feature bank, and uses it for same-anchor replacement. The selected points are combined with the unchanged input points before downstream anomaly detection.}
    \label{fig:pipeline}
\end{figure*}

$\mathcal{PC}^{2}$-AD compensates sparse test point clouds using the geometric information available in normal training data. As shown in Fig.~\ref{fig:pipeline}, TCG first adapts a distance estimator to the target domain and generates a dense candidate pool. GACF selects candidates that provide spatial coverage around the original observations. NPPC then refines this selection according to local normality consistency with input anchors. The selected points are combined with the unchanged input points before downstream anomaly detection. TDA is the training stage of TCG, while Contrastive Normality Pretraining (CNP) and Mean-Shifted Contrastive Fine-Tuning (MSCF) train the representation used by NPPC for same-anchor replacement.

\subsection{Target Domain Candidate Generation}

High-resolution normal training clouds provide geometric information that is less densely sampled by low-resolution test sensors. TCG uses this information to generate candidate points for sparse test clouds. It adapts a pretrained upsampler to the target domain and constructs a dense pool for subsequent geometric filtering and normality-guided selection.

Given a generic point cloud upsampling model $\mathcal{R}_{\phi}$, we formulate TCG as two stages: Target Domain Adaptation (TDA) during training and Candidate Generation (CG) during inference. In this work, we instantiate $\mathcal{R}_{\phi}$ with Grad-PU~\cite{He_2023_CVPR}, whose pretrained distance estimator is denoted by $D_{\theta}$.

Existing point cloud upsampling datasets and 3D anomaly detection datasets belong to different domains, so $D_{\theta}$ may not accurately characterize the normal local geometry of the target domain. During training, \textbf{Target Domain Adaptation (TDA)} adapts $D_{\theta}$ using only $\mathcal{P}^{\text{train}}$. Specifically, we construct a low-resolution counterpart for each normal training point cloud $\mathbf{P}_i^{\text{train}}$ as
\begin{equation}
\mathbf{P}_i^{\text{LR}}
=
\text{FPS}_{r}
\left(\mathbf{P}_i^{\text{train}}\right),
\label{eq:tda_downsampling}
\end{equation}
where $i=1,\ldots,N$ and $\text{FPS}_{r}$ retains $1/r$ of the input points. The original $\mathbf{P}_i^{\text{train}}$ serves as the corresponding high-resolution normal reference. Initial candidate points are generated from $\mathbf{P}_i^{\text{LR}}$ through midpoint interpolation.

We perturb the interpolated candidates with Gaussian noise to construct query samples around their initial locations. The queries and corresponding high-resolution normal references are used to fine-tune $D_{\theta}$. The adapted distance estimator $D_{\theta^\star}$ is then frozen.

During inference, \textbf{Candidate Generation (CG)} applies midpoint interpolation with ratio $r$ to each $\mathbf{P}_i^{\text{test}}\in\mathcal{P}^{\text{test}}$ and obtains an initial candidate set $C_i^{(0)}$. Since midpoint interpolation does not constrain the generated points to the target-domain normal surface, their coordinates are further refined using the frozen $D_{\theta^\star}$. For the $q$-th candidate point $x_{i,q}^{(t)}$, the coordinate update is defined as
\begin{equation}
x_{i,q}^{(t+1)}
=
x_{i,q}^{(t)}
-
\eta
\nabla_{x_{i,q}^{(t)}}
D_{\theta^\star}
\left(x_{i,q}^{(t)}\right),
\label{eq:cg_refinement}
\end{equation}
where $\eta$ denotes the step size and the parameters of $D_{\theta^\star}$ remain fixed. After $T$ refinement iterations, a deterministic blockwise FPS scheme based on Morton ordering (Morton-block FPS) is applied to the refined points to obtain the candidate pool $C_i$ for subsequent geometric filtering and normality-guided point compensation.

\subsection{Geometry-Aware Candidate Filtering}

Reliable resolution compensation requires the added points to maintain appropriate spacing from the original observations and provide balanced spatial coverage. However, the candidate pool $C_i$ generated by TCG may still contain redundant, geometrically unreliable, or spatially clustered points. We therefore introduce Geometry-Aware Candidate Filtering (GACF). It first performs scale-adaptive filtering according to the relative distance between each candidate and its nearest input anchor, followed by region-budgeted voxel sampling to promote balanced spatial coverage. The selected candidates form $G_i$, while the remaining geometrically valid candidates form $U_i$.
An effective compensation point should provide additional spatial coverage without deviating excessively from the observed geometry. Given a low-resolution test point cloud $\mathbf{P}_i^{\text{test}}$ and its candidate pool $C_i$, GACF estimates the sampling spacing as
\begin{equation}
\rho_i
=
\frac{1}{n_i^{\text{test}}}
\sum_{p\in\mathbf{P}_i^{\text{test}}}
\min_{p'\in\mathbf{P}_i^{\text{test}},\,p'\neq p}
\left\|p-p'\right\|_2.
\label{eq:gacf_spacing}
\end{equation}
Herein, the nearest input point of each candidate $x\in C_i$ is treated as its anchor and defined as
\begin{equation}
p_x
=
\arg\min_{p\in\mathbf{P}_i^{\text{test}}}
\left\|x-p\right\|_2.
\label{eq:gacf_anchor}
\end{equation}
The distance $\|x-p_x\|_2$ measures the relative displacement between the candidate and its anchor. A small anchor distance provides limited additional coverage, whereas an excessively large distance indicates low geometric reliability.

Let $T_i$ and $L_i$ denote the candidate sets retained by strict and relaxed anchor-distance filtering, respectively. Candidates retained by these filters may still be spatially clustered, limiting their contribution to surface coverage. To suppress this redundancy, GACF partitions the 3D space into cubic voxels with side length $\delta\rho_i$ and retains at most one candidate in each occupied voxel, prioritizing $T_i$ over $L_i$.

Voxel-level deduplication controls local redundancy but does not ensure coverage across the entire point cloud. GACF therefore allocates candidates across larger spatial regions with side length $\gamma\rho_i$. Each candidate is assigned to the region containing its anchor. For a target output size $n_i^{\mathrm{target}}$, let $M_i=n_i^{\mathrm{target}}-n_i^{\mathrm{test}}$ be the compensation budget. The target quota of region $R$ is
\begin{equation}
q_{i,R}
=
M_i
\frac{n_{i,R}^{\text{test}}}
{n_i^{\text{test}}},
\label{eq:gacf_budget}
\end{equation}
where $n_{i,R}^{\text{test}}$ is the number of input points in region $R$. The fractional quotas are converted into integer budgets using the largest-remainder rule, ensuring that the total allocation equals $M_i$. Within each region, candidates from $T_i$ are selected before those from $L_i$. Candidates from neighboring regions and then the remaining global candidates are used when the local quota cannot be filled. The selected candidates form $G_i$, while the remaining geometrically valid candidates form $U_i$.

\subsection{Normality-Preserving Point Compensation}

Resolution compensation should increase sampling density while maintaining the local anomaly evidence carried by the original observations. Newly added points may obscure genuine anomalies or introduce spurious responses by changing local neighborhoods. NPPC therefore selects compensation points according to their local normality relative to input anchors. Given the normal training set $\mathcal{P}^{\text{train}}$ and the candidate sets $G_i$ and $U_i$ for $\mathbf{P}_i^{\text{test}}$, it learns local normality representations and uses them for same-anchor one-to-one replacement. This operation improves candidate--anchor consistency while preserving the number of selected points assigned to each anchor. The resulting set $S_i$ is combined with the original observations.
To align normality learning with the candidate distribution encountered during inference, each normal cloud $\mathbf{P}_i^{\text{train}}$ is downsampled and processed by CG using the frozen $D_{\theta^\star}$:
\begin{equation}
C_i^{\text{train}}
=
\text{CG}
\left(
\text{FPS}_{r}
\left(\mathbf{P}_i^{\text{train}}\right);
D_{\theta^\star}
\right).
\label{eq:nppc_training_candidates}
\end{equation}
Local centers are subsequently selected from $C_i^{\text{train}}$ by FPS. Given the $\ell$-th center $z_{i,\ell}$, the corresponding normal patch is constructed as
\begin{equation}
Q_{i,\ell}^{\text{normal}}
=
\text{KNN}_{k}
\left(
z_{i,\ell},
C_i^{\text{train}}
\right).
\label{eq:nppc_normal_patch}
\end{equation}

For each $Q_{i,\ell}^{\text{normal}}$, two views $V_{i,\ell}^{(1)}$ and $V_{i,\ell}^{(2)}$ are independently generated through rotation, scaling, coordinate jittering, and random point dropout. Both views are encoded using a shared PointNet~\cite{qi2017pointnet}, followed by feature normalization:
\begin{equation}
\mathbf{F}_{i,\ell}^{(v)}
=
\text{Normalize}
\left(
\text{PointNet}
\left(V_{i,\ell}^{(v)}\right)
\right),
\label{eq:nppc_pointnet_feature}
\end{equation}
where $v\in\{1,2\}$. NPPC first learns augmentation-invariant representations by treating the two views of the same patch as a positive pair and views from different patches as negative pairs. It then averages the normalized features of all normal patches from category $c$ to obtain the category-specific feature center $\boldsymbol{\mu}_c$. Each augmented feature is shifted by subtracting $\boldsymbol{\mu}_c$ and normalized again to refine the representation relative to the category-specific normal pattern.

After refinement, the resulting feature extractor is frozen and denoted by $E$. NPPC uses $E$ to re-extract normalized features from all normal training patches of category $c$, forming the category-specific normal feature bank $F_c$.
Given $\mathbf{P}_i^{\text{test}}$, its complete candidate pool $C_i$, and the candidate sets $G_i$ and $U_i$, NPPC evaluates every point $q\in\mathbf{P}_i^{\text{test}}\cup G_i\cup U_i$ under the shared neighborhood context $\mathbf{P}_i^{\text{test}}\cup C_i$. The $k$ nearest neighbors of $q$ form a local patch $Q_i(q)$. After centering and scale normalization, its local feature is extracted as
\begin{equation}
\mathbf{F}_i(q)
=
\text{Normalize}
\left(
E\left(Q_i(q)\right)
\right).
\label{eq:nppc_local_feature}
\end{equation}

Let $\{\mathbf{F}_{c,h}(q)\}_{h=1}^{K_{\text{nn}}}$ denote the $K_{\text{nn}}$ most similar features to $\mathbf{F}_i(q)$ retrieved from $F_c$. The local normality distance of $q$ is defined as
\begin{equation}
d_i^{\text{normal}}(q)
=
\sum_{h=1}^{K_{\text{nn}}}
\left[
1-
\mathbf{F}_i(q)
\cdot
\mathbf{F}_{c,h}(q)
\right].
\label{eq:nppc_normality_distance}
\end{equation}
A smaller $d_i^{\text{normal}}(q)$ indicates stronger consistency with the learned normal patterns.

NPPC does not directly favor candidates with the smallest normality distance. It favors candidates whose normality level is close to that of their input anchor, with the aim of reducing the risk of suppressing anomaly evidence. This criterion retains the anchor as the local reference even when its normality distance is large. All evaluated points are ranked in descending order of $d_i^{\text{normal}}$, and $\pi_i(q)$ denotes the rank of $q$. If a GACF-selected candidate is already among the closest same-anchor neighbors of an anchor in the normality ranking, the current selection is retained without replacement. For a candidate $q$ with anchor $p_q$, its rank gap and normality-distance gap are $g_i(q)=|\pi_i(q)-\pi_i(p_q)|$ and $h_i(q)=|d_i^{\text{normal}}(q)-d_i^{\text{normal}}(p_q)|$, respectively.
For each anchor, a candidate from $U_i$ with a smaller rank gap is selected as the incoming candidate $q^{\text{in}}$, while a candidate from $G_i$ with a larger rank gap is selected as the outgoing candidate $q^{\text{out}}$. A proposed replacement must satisfy
\begin{equation}
g_i\left(q^{\text{in}}\right)+\kappa
\leq
g_i\left(q^{\text{out}}\right)
\;\text{and}\;
h_i\left(q^{\text{in}}\right)
\leq
h_i\left(q^{\text{out}}\right),
\label{eq:nppc_replacement}
\end{equation}
where $\kappa$ specifies the minimum required improvement in rank alignment. Ambiguous pairs with larger normality gaps are further screened using stricter confidence-dependent margins. As the normality gap increases, a stronger rank improvement is required, while candidates exceeding the admissible gap are directly rejected.

Accepted replacements exchange candidates sharing the same anchor. Each candidate participates in at most one replacement, so the number of selected candidates and their allocation across anchors remain unchanged. The resulting compensation points form $S_i$.

\subsection{Training and Inference}
\label{sec:training_inference}

\textbf{Training.}
Given $\mathcal{P}^{\text{train}}$, $\mathcal{PC}^{2}$-AD is trained using only normal samples through three sequential stages.

\textit{Stage 1: Target Domain Adaptation (TDA).}
The pretrained distance estimator $D_{\theta}$ is first adapted to the target-domain geometry. Let $x_b$ denote a perturbed query point and $d_b$ denote its nearest distance to the corresponding high-resolution normal reference. Given $B_d$ query points, the adaptation loss is defined as
\begin{equation}
\mathcal{L}_{\text{TDA}}
=
\frac{1}{B_d}
\sum_{b=1}^{B_d}
\left|
D_{\theta}(x_b)
-
d_b
\right|.
\label{eq:tda_loss}
\end{equation}
After this stage, the resulting distance estimator is frozen and denoted by $D_{\theta^\star}$. The frozen $D_{\theta^\star}$ is then used to construct $C_i^{\text{train}}$ and $Q_{i,\ell}^{\text{normal}}$ without updating model parameters.

\textit{Stage 2: Contrastive Normality Pretraining (CNP).}
NPPC is subsequently optimized using the normal patches defined in Eq.~\eqref{eq:nppc_normal_patch}. For two normalized features $\mathbf{F}$ and $\mathbf{F}'$, their cosine similarity is denoted by $\text{cos}(\mathbf{F},\mathbf{F}')$, and $\tau$ is the temperature parameter. Let $\mathbf{F}_b^{(1)}$ and $\mathbf{F}_b^{(2)}$ denote the two features of the $b$-th patch in a batch of size $B$. The CNP loss follows a symmetric InfoNCE formulation~\cite{oord2018representation} and is defined as
\begin{equation}
\begin{aligned}
\mathcal{L}_{\text{CNP}}
=
-\frac{1}{2B}
\sum_{b=1}^{B}
\Bigg[
&\log
\frac{
\exp\left(\text{cos}\left(\mathbf{F}_b^{(1)},\mathbf{F}_b^{(2)}\right)/\tau\right)
}{
\displaystyle\sum_{a=1}^{B}
\exp\left(\text{cos}\left(\mathbf{F}_b^{(1)},\mathbf{F}_a^{(2)}\right)/\tau\right)
}\\
+
&\log
\frac{
\exp\left(\text{cos}\left(\mathbf{F}_b^{(2)},\mathbf{F}_b^{(1)}\right)/\tau\right)
}{
\displaystyle\sum_{a=1}^{B}
\exp\left(\text{cos}\left(\mathbf{F}_b^{(2)},\mathbf{F}_a^{(1)}\right)/\tau\right)
}
\Bigg].
\end{aligned}
\label{eq:nppc_contrastive_loss}
\end{equation}

After CNP, a checkpoint is selected to re-extract normalized features from all normal patches of category $c$. Their average forms $\boldsymbol{\mu}_c$, which is not normalized again. Computing $\boldsymbol{\mu}_c$ does not update model parameters.

\textit{Stage 3: Mean-Shifted Contrastive Fine-Tuning (MSCF).}
MSCF uses a mean-shifted contrastive objective~\cite{reiss2023mean}, built on a two-view contrastive formulation~\cite{chen2020simple}, for normal point cloud patches and is initialized from the same checkpoint selected after CNP. Let $\{\mathbf{F}_u^{\text{shift}}\}_{u=1}^{2B}$ denote all features obtained by subtracting $\boldsymbol{\mu}_c$ and normalizing again, and let $\mathbf{F}_{u^+}^{\text{shift}}$ denote the other view of the same patch. The MSCF loss is
\begin{equation}
\mathcal{L}_{\text{MSCF}}
=
-\frac{1}{2B}
\sum_{u=1}^{2B}
\log
\frac{
\exp\left(
\text{cos}\left(
\mathbf{F}_u^{\text{shift}},
\mathbf{F}_{u^+}^{\text{shift}}
\right)/\tau
\right)
}{
\displaystyle
\sum_{\substack{k=1\\k\ne u}}^{2B}
\exp\left(
\text{cos}\left(
\mathbf{F}_u^{\text{shift}},
\mathbf{F}_k^{\text{shift}}
\right)/\tau
\right)
}.
\label{eq:nppc_shifted_loss}
\end{equation}
After MSCF, the resulting feature extractor is frozen and denoted by $E$. The normal feature bank $F_c$ is then constructed without further optimization. Thus, $\mathcal{L}_{\text{TDA}}$, $\mathcal{L}_{\text{CNP}}$, and $\mathcal{L}_{\text{MSCF}}$ are optimized sequentially rather than combined into a joint objective.

\textbf{Inference.}
Given a low-resolution test point cloud $\mathbf{P}_i^{\text{test}}$, CG first uses the frozen $D_{\theta^\star}$ to generate $C_i$. GACF then selects geometrically reliable and spatially distributed candidates, producing $G_i$ and $U_i$. NPPC uses the frozen $E$ and $F_c$ to perform same-anchor one-to-one replacement and obtain $S_i$. CG updates only candidate coordinates, and no model parameter is optimized during inference.
The resolution-compensated test point cloud is obtained as
\begin{equation}
\widehat{\mathbf{P}}_i^{\text{test}}
=
\mathbf{P}_i^{\text{test}}
\cup
S_i.
\label{eq:resolution_compensation}
\end{equation}
Applying this procedure to all test point clouds yields $\widehat{\mathcal{P}}^{\text{test}}=\{\widehat{\mathbf{P}}_i^{\text{test}}\}_{i=1}^{m}$, which is subsequently passed to a downstream 3D anomaly detector.

\section{Experiments}

\subsection{Datasets}

We evaluate $\mathcal{PC}^{2}$-AD on two 3D anomaly detection benchmarks, Anomaly-ShapeNet~\cite{anomaly_shapenet} and Real3D-AD~\cite{liu2023neurips-real3dad}, to investigate resolution compensation for sparse test point clouds while keeping the normal training data fixed. \textbf{(1) Anomaly-ShapeNet.} We evaluate Anomaly-ShapeNet and Anomaly-ShapeNet-new, which contain 1,312 test samples from 40 categories and 411 test samples from 12 categories, respectively. Each category contains four normal training samples. The average point count of the normal training clouds in each category is used as the resolution reference. Test clouds are downsampled by FPS to an integer point count obtained by dividing this reference by four and rounding down, and the target point count is set to four times the resulting input point count. Thus, the resolution-compensated outputs closely match the normal training resolution of each category. \textbf{(2) Real3D-AD.} Real3D-AD contains 1,254 samples from 12 categories, including 48 normal training samples and 1,206 test samples. For each original test cloud with $n_j^{\mathrm{orig}}$ points, FPS constructs a low-resolution input with $\lfloor n_j^{\mathrm{orig}}/4 \rfloor$ points, which is then processed by $\mathcal{PC}^{2}$-AD to obtain a resolution-compensated output with $n_j^{\mathrm{orig}}$ points. This setting evaluates the effect of restoring the original test point count after controlled downsampling.

\subsection{Evaluation Protocol}

Point coordinates and point labels are downsampled using the same FPS indices to construct the low-resolution inputs, which are then processed by $\mathcal{PC}^{2}$-AD to obtain the resolution-compensated outputs. For each downstream detector, only the test point cloud representation changes, while all other training, inference, and evaluation settings remain unchanged. Object-level evaluation uses all points. For point-level evaluation, the added compensation points participate in inference, while metrics are computed only on the original input points. For the main downstream comparisons, we report O-AUROC and P-AUROC. O-AUPRC and P-AUPRC are additionally used in the aggregate metrics for the ablation and parameter-sensitivity analyses. Object-Avg denotes the average of O-AUROC and O-AUPRC, while Point-Avg denotes the average of P-AUROC and P-AUPRC. Avg.\ AUROC denotes the average of O-AUROC and P-AUROC, while Avg.\ AUPRC denotes the average of O-AUPRC and P-AUPRC. All category-level averages are computed as unweighted macro averages across categories. In the quantitative tables, baseline rows denote evaluation on the low-resolution inputs, while rows with the $\mathcal{PC}^{2}$-AD subscript denote evaluation on the corresponding resolution-compensated outputs. Red indicates that $\mathcal{PC}^{2}$-AD yields a higher value than the corresponding baseline, while blue indicates that the baseline yields a higher value.

\subsection{Baselines}

We evaluate $\mathcal{PC}^{2}$-AD with six downstream 3D anomaly detection baselines: PatchCore~\cite{roth2022patchcore}, MC3D-AD~\cite{cheng2025mc3dad}, Reg2Inv~\cite{yu2025reg2inv}, BTF~\cite{horwitz2023btf}, M3DM~\cite{wang2023cvpr-multimodal}, and Point-BERT~\cite{yu2022pointbert}. PatchCore and BTF use geometric features based on Fast Point Feature Histograms (FPFH)~\cite{rusu2009fpfh}, MC3D-AD performs feature reconstruction, and Reg2Inv combines registration with memory-based anomaly detection. M3DM uses its Point-MAE-based 3D branch, while Point-BERT constructs a normal feature bank from pretrained Transformer representations.

\subsection{Implementation Details}

All experiments are conducted on a server equipped with five NVIDIA GeForce RTX 4090 D GPUs. $\mathcal{PC}^{2}$-AD is implemented in Python 3.8 with PyTorch 2.0.1 and CUDA 11.8. During TDA, the distance estimator is optimized for 90 epochs using Adam. NPPC adopts a PointNet encoder with a 128-dimensional projection space. Its CNP and MSCF training stages run for 60 and 50 epochs, respectively, with MSCF initialized from the epoch-50 checkpoint of CNP. The local patch size is set to 1,024 points. During inference, the query coordinates are refined for 10 iterations along the negative gradient of the distance field with a step size of 500. All main experiments use the same default $\mathcal{PC}^{2}$-AD configuration. Parameter sensitivity varies one parameter setting at a time, with jointly constrained parameter pairs varied together; 12 core parameters are reported in the main paper and the remaining 17 are provided in the supplementary material. PatchCore results are averaged over five repeated runs on Anomaly-ShapeNet and three repeated runs on Anomaly-ShapeNet-new and Real3D-AD. The other baselines follow a paired evaluation protocol in which only the test point cloud representation changes while all other settings remain fixed.

\subsection{Quantitative Results}

\subsubsection{Results on Anomaly-ShapeNet}
Table~\ref{tab:asn_pcd_results} compares six downstream detectors before and after resolution compensation. All six improve in Avg.\ AUROC, with the largest gains from BTF and PatchCore at 10.8 and 8.7 percentage points, respectively. For PatchCore, this aggregate gain combines an increase in P-AUROC from 56.6\% to 77.3\% with a decrease in O-AUROC from 62.5\% to 59.2\%.

\subsubsection{Results on Anomaly-ShapeNet-new}
On Anomaly-ShapeNet-new, all six detectors improve in Avg.\ AUROC (Table~\ref{tab:asn_new_pcd_results}). BTF and PatchCore gain the most at 12.4 and 8.6 percentage points, respectively. PatchCore's P-AUROC increases from 70.7\% to 88.2\%, while its O-AUROC decreases from 60.4\% to 60.0\%. 

\begin{table*}[!t] 
\centering 

\caption{Per-category O-AUROC/P-AUROC~($\uparrow$) results (\%) on Anomaly-ShapeNet. Values in parentheses give the change in Avg.\ AUROC, the mean of O-AUROC and P-AUROC, in percentage points.} 
\label{tab:asn_pcd_results}

\begin{adjustbox}{max width=\textwidth} 
\begin{tabular}{l|cccccccccccccr} 
\hline 
\multicolumn{15}{c}{\textbf{O-AUROC/P-AUROC}} \\ 
\hline 
\textbf{Method} & \textbf{ashtray0} & \textbf{bag0} & \textbf{bottle0} & \textbf{bottle1} & \textbf{bottle3} & \textbf{bowl0} & \textbf{bowl1} & \textbf{bowl2} & \textbf{bowl3} & \textbf{bowl4} & \textbf{bowl5} & \textbf{bucket0} & \textbf{bucket1} & \multicolumn{1}{c}{\textbf{cap0}} \\ 
\hline 
 
PatchCore & 64.8/39.9 & 53.2/82.6 & \textcolor{baselineblue}{85.7}/81.1 & \textcolor{baselineblue}{89.2}/\textcolor{baselineblue}{78.1} & \textcolor{baselineblue}{84.2}/76.9 & \textcolor{baselineblue}{93.7}/90.5 & \textcolor{baselineblue}{77.5}/41.9 & 60.1/70.0 & 45.4/63.0 & \textcolor{baselineblue}{54.7}/58.8 & 33.6/41.8 & \textcolor{baselineblue}{70.9}/67.3 & 53.4/47.7 & 41.6/25.6 \\ 
MC3D-AD & 51.0/57.4 & \textcolor{baselineblue}{62.9}/\textcolor{baselineblue}{60.7} & \textcolor{baselineblue}{72.4}/\textcolor{baselineblue}{62.7} & \textcolor{baselineblue}{55.8}/45.8 & \textcolor{baselineblue}{57.8}/52.5 & \textcolor{baselineblue}{59.6}/55.3 & \textcolor{baselineblue}{60.7}/54.5 & \textcolor{baselineblue}{64.8}/\textcolor{baselineblue}{57.7} & 60.4/53.6 & \textcolor{baselineblue}{44.1}/\textcolor{baselineblue}{49.2} & 39.6/\textcolor{baselineblue}{48.8} & \textcolor{baselineblue}{48.6}/\textcolor{baselineblue}{55.0} & 54.0/52.5 & 47.0/54.0 \\ 
Reg2Inv & 91.0/85.4 & 100.0/99.7 & 100.0/99.6 & 100.0/87.9 & 100.0/\textcolor{baselineblue}{90.1} & 100.0/99.5 & 84.8/90.4 & 82.6/\textcolor{baselineblue}{95.1} & 71.1/90.9 & 75.6/80.5 & \textcolor{baselineblue}{67.7}/\textcolor{baselineblue}{86.5} & \textcolor{baselineblue}{94.9}/77.7 & 80.0/\textcolor{baselineblue}{93.0} & 90.4/93.3 \\ 
BTF & \textcolor{baselineblue}{62.4}/42.5 & 35.7/56.3 & 64.8/64.7 & 66.7/\textcolor{baselineblue}{72.6} & 50.8/59.7 & 50.0/69.2 & \textcolor{baselineblue}{61.9}/42.9 & 57.4/65.3 & 61.9/55.2 & 41.9/57.6 & 48.8/42.0 & 55.6/67.4 & 44.1/56.8 & 58.5/33.0 \\ 
M3DM & 55.2/57.3 & 51.4/60.2 & \textcolor{baselineblue}{51.0}/57.2 & \textcolor{baselineblue}{51.6}/58.7 & 51.1/71.2 & 48.5/59.4 & \textcolor{baselineblue}{52.6}/35.6 & \textcolor{baselineblue}{51.1}/40.6 & \textcolor{baselineblue}{43.7}/34.8 & \textcolor{baselineblue}{49.3}/56.8 & \textcolor{baselineblue}{53.7}/\textcolor{baselineblue}{51.2} & \textcolor{baselineblue}{46.3}/44.0 & \textcolor{baselineblue}{46.3}/62.9 & \textcolor{baselineblue}{46.7}/57.0 \\ 
Point-BERT & \textcolor{baselineblue}{69.0}/62.8 & 30.0/43.3 & 32.9/55.3 & \textcolor{baselineblue}{54.4}/\textcolor{baselineblue}{67.0} & 40.6/78.8 & 47.0/52.4 & 40.0/\textcolor{baselineblue}{44.4} & 47.4/55.8 & \textcolor{baselineblue}{49.3}/\textcolor{baselineblue}{58.8} & \textcolor{baselineblue}{50.0}/\textcolor{baselineblue}{56.0} & 46.0/47.8 & \textcolor{baselineblue}{51.1}/45.2 & 36.2/51.7 & 51.9/60.5 \\ 
\rowcolor{gray!8}PatchCore$_{\mathcal{PC}^{2}\text{-}AD}$ & \textcolor{darkred}{73.0}/\textcolor{darkred}{55.3} & \textcolor{darkred}{65.1}/\textcolor{darkred}{87.7} & 76.3/81.1 & 58.7/71.9 & 71.9/\textcolor{darkred}{89.5} & 83.5/\textcolor{darkred}{93.3} & 51.6/\textcolor{darkred}{68.9} & \textcolor{darkred}{62.1}/\textcolor{darkred}{90.1} & \textcolor{darkred}{70.5}/\textcolor{darkred}{96.5} & 45.8/\textcolor{darkred}{91.8} & \textcolor{darkred}{47.7}/\textcolor{darkred}{66.3} & 39.9/\textcolor{darkred}{74.2} & \textcolor{darkred}{55.4}/\textcolor{darkred}{80.3} & \textcolor{darkred}{60.6}/\textcolor{darkred}{96.4} \\ 
\rowcolor{gray!8}MC3D-AD$_{\mathcal{PC}^{2}\text{-}AD}$ & \textcolor{darkred}{75.7}/\textcolor{darkred}{69.6} & 40.5/47.7 & 57.1/50.4 & 54.4/\textcolor{darkred}{61.8} & 46.3/\textcolor{darkred}{55.1} & 52.2/\textcolor{darkred}{58.0} & 58.9/\textcolor{darkred}{58.8} & 56.3/49.9 & 60.4/\textcolor{darkred}{59.5} & 40.0/45.6 & \textcolor{darkred}{46.7}/47.7 & 43.8/51.1 & \textcolor{darkred}{76.2}/\textcolor{darkred}{57.9} & \textcolor{darkred}{63.7}/\textcolor{darkred}{68.6} \\ 
\rowcolor{gray!8}Reg2Inv$_{\mathcal{PC}^{2}\text{-}AD}$ & \textcolor{darkred}{99.0}/\textcolor{darkred}{88.4} & 100.0/99.7 & 100.0/99.6 & 100.0/\textcolor{darkred}{88.2} & 100.0/90.0 & 100.0/99.5 & \textcolor{darkred}{85.9}/\textcolor{darkred}{91.0} & \textcolor{darkred}{87.8}/95.0 & \textcolor{darkred}{94.8}/\textcolor{darkred}{97.7} & \textcolor{darkred}{83.7}/\textcolor{darkred}{84.3} & 35.4/86.3 & 91.7/\textcolor{darkred}{79.5} & \textcolor{darkred}{86.3}/88.8 & \textcolor{darkred}{94.8}/\textcolor{darkred}{93.8} \\ 
\rowcolor{gray!8}BTF$_{\mathcal{PC}^{2}\text{-}AD}$ & 60.0/\textcolor{darkred}{59.8} & \textcolor{darkred}{44.8}/\textcolor{darkred}{81.8} & 64.8/\textcolor{darkred}{79.5} & \textcolor{darkred}{73.7}/70.2 & \textcolor{darkred}{69.2}/\textcolor{darkred}{84.5} & \textcolor{darkred}{83.0}/\textcolor{darkred}{88.5} & 44.1/\textcolor{darkred}{60.2} & \textcolor{darkred}{71.1}/\textcolor{darkred}{80.6} & \textcolor{darkred}{74.4}/\textcolor{darkred}{83.4} & \textcolor{darkred}{64.1}/\textcolor{darkred}{70.9} & \textcolor{darkred}{57.9}/\textcolor{darkred}{64.6} & \textcolor{darkred}{56.2}/\textcolor{darkred}{71.7} & \textcolor{darkred}{51.7}/\textcolor{darkred}{71.2} & \textcolor{darkred}{74.4}/\textcolor{darkred}{91.8} \\ 
\rowcolor{gray!8}M3DM$_{\mathcal{PC}^{2}\text{-}AD}$ & \textcolor{darkred}{59.0}/\textcolor{darkred}{63.0} & \textcolor{darkred}{53.8}/\textcolor{darkred}{68.2} & 47.6/\textcolor{darkred}{62.5} & 48.8/\textcolor{darkred}{61.2} & \textcolor{darkred}{56.5}/\textcolor{darkred}{80.1} & \textcolor{darkred}{51.5}/\textcolor{darkred}{65.9} & 47.8/\textcolor{darkred}{43.4} & 45.6/\textcolor{darkred}{49.0} & 41.9/\textcolor{darkred}{52.6} & 48.5/\textcolor{darkred}{60.5} & 46.3/42.1 & 44.4/\textcolor{darkred}{52.7} & 44.4/\textcolor{darkred}{64.7} & 44.4/\textcolor{darkred}{69.4} \\ 
\rowcolor{gray!8}Point-BERT$_{\mathcal{PC}^{2}\text{-}AD}$ & 47.6/\textcolor{darkred}{63.6} & \textcolor{darkred}{50.0}/\textcolor{darkred}{51.9} & \textcolor{darkred}{51.4}/\textcolor{darkred}{62.2} & 53.0/61.7 & \textcolor{darkred}{49.8}/\textcolor{darkred}{82.6} & \textcolor{darkred}{60.4}/\textcolor{darkred}{63.4} & \textcolor{darkred}{52.6}/42.8 & \textcolor{darkred}{49.6}/\textcolor{darkred}{57.4} & 39.3/55.1 & 43.7/54.4 & \textcolor{darkred}{69.8}/\textcolor{darkred}{58.8} & 41.6/\textcolor{darkred}{51.7} & \textcolor{darkred}{44.8}/\textcolor{darkred}{60.0} & \textcolor{darkred}{54.1}/\textcolor{darkred}{69.4} \\ 
\hline 
\hline 
 
\textbf{Method} & \textbf{cap3} & \textbf{cap4} & \textbf{cap5} & \textbf{cup0} & \textbf{cup1} & \textbf{eraser0} & \textbf{headset0} & \textbf{headset1} & \textbf{helmet0} & \textbf{helmet1} & \textbf{helmet2} & \textbf{helmet3} & \textbf{jar0} & \multicolumn{1}{c}{\textbf{microphone0}} \\ 
\hline 
 
PatchCore & 60.8/46.3 & \textcolor{baselineblue}{63.9}/35.8 & \textcolor{baselineblue}{78.8}/37.3 & 47.5/54.2 & \textcolor{baselineblue}{59.7}/35.8 & 73.7/89.6 & 60.4/34.4 & \textcolor{baselineblue}{73.0}/67.5 & 50.0/54.3 & \textcolor{baselineblue}{55.6}/44.0 & \textcolor{baselineblue}{66.3}/65.2 & \textcolor{baselineblue}{51.6}/70.9 & 62.4/72.1 & \textcolor{baselineblue}{99.2}/\textcolor{baselineblue}{92.4} \\ 
MC3D-AD & 42.1/47.1 & 50.5/\textcolor{baselineblue}{52.5} & 36.1/36.7 & \textcolor{baselineblue}{57.1}/\textcolor{baselineblue}{59.4} & 60.0/57.4 & 44.8/48.3 & 41.3/48.0 & 46.2/46.7 & \textcolor{baselineblue}{51.9}/47.2 & \textcolor{baselineblue}{60.0}/59.6 & \textcolor{baselineblue}{47.0}/48.3 & \textcolor{baselineblue}{54.8}/\textcolor{baselineblue}{58.5} & \textcolor{baselineblue}{61.4}/54.2 & \textcolor{baselineblue}{39.0}/\textcolor{baselineblue}{51.5} \\ 
Reg2Inv & 83.9/\textcolor{baselineblue}{95.1} & \textcolor{baselineblue}{68.8}/\textcolor{baselineblue}{92.6} & \textcolor{baselineblue}{88.4}/\textcolor{baselineblue}{98.1} & \textcolor{baselineblue}{82.9}/90.4 & \textcolor{baselineblue}{92.9}/\textcolor{baselineblue}{93.9} & 100.0/99.6 & 88.0/92.0 & 45.2/91.8 & \textcolor{baselineblue}{80.6}/93.4 & \textcolor{baselineblue}{100.0}/\textcolor{baselineblue}{95.4} & \textcolor{baselineblue}{89.3}/91.6 & \textcolor{baselineblue}{92.4}/97.8 & 100.0/99.6 & 100.0/99.4 \\ 
BTF & 43.2/41.7 & \textcolor{baselineblue}{53.7}/45.9 & \textcolor{baselineblue}{82.1}/42.3 & \textcolor{baselineblue}{47.6}/56.2 & \textcolor{baselineblue}{58.6}/36.4 & 61.0/72.6 & 63.1/43.0 & 47.6/62.4 & \textcolor{baselineblue}{63.8}/54.2 & \textcolor{baselineblue}{80.5}/45.9 & 53.3/52.4 & 48.8/59.3 & \textcolor{baselineblue}{75.2}/62.4 & 51.0/76.4 \\ 
M3DM & 43.5/57.8 & \textcolor{baselineblue}{63.2}/55.1 & \textcolor{baselineblue}{57.9}/\textcolor{baselineblue}{61.3} & 57.1/46.5 & \textcolor{baselineblue}{55.7}/49.2 & 52.9/36.5 & 48.0/50.4 & \textcolor{baselineblue}{51.9}/40.1 & 47.0/55.0 & 49.0/50.2 & \textcolor{baselineblue}{50.1}/63.6 & \textcolor{baselineblue}{54.5}/38.0 & 45.2/66.7 & \textcolor{baselineblue}{57.6}/\textcolor{baselineblue}{65.2} \\ 
Point-BERT & \textcolor{baselineblue}{64.2}/\textcolor{baselineblue}{66.1} & \textcolor{baselineblue}{56.8}/\textcolor{baselineblue}{64.9} & 63.9/54.1 & \textcolor{baselineblue}{61.4}/56.5 & 51.9/50.6 & \textcolor{baselineblue}{54.3}/\textcolor{baselineblue}{52.4} & 48.4/54.1 & 51.4/50.1 & 50.4/57.1 & \textcolor{baselineblue}{70.0}/47.8 & \textcolor{baselineblue}{48.7}/\textcolor{baselineblue}{58.6} & 50.3/48.8 & \textcolor{baselineblue}{65.2}/\textcolor{baselineblue}{68.8} & \textcolor{baselineblue}{54.3}/\textcolor{baselineblue}{61.8} \\ 
\rowcolor{gray!8}PatchCore$_{\mathcal{PC}^{2}\text{-}AD}$ & \textcolor{darkred}{65.8}/\textcolor{darkred}{77.0} & 44.1/\textcolor{darkred}{70.5} & 54.9/\textcolor{darkred}{59.2} & \textcolor{darkred}{57.2}/\textcolor{darkred}{91.7} & 55.8/\textcolor{darkred}{55.9} & \textcolor{darkred}{92.3}/\textcolor{darkred}{92.8} & \textcolor{darkred}{71.6}/\textcolor{darkred}{61.5} & 51.3/\textcolor{darkred}{84.3} & \textcolor{darkred}{54.4}/\textcolor{darkred}{79.0} & 38.9/\textcolor{darkred}{54.4} & 65.3/\textcolor{darkred}{92.1} & 46.1/\textcolor{darkred}{75.5} & \textcolor{darkred}{70.0}/\textcolor{darkred}{94.5} & 74.2/85.6 \\ 
\rowcolor{gray!8}MC3D-AD$_{\mathcal{PC}^{2}\text{-}AD}$ & \textcolor{darkred}{51.9}/\textcolor{darkred}{51.0} & \textcolor{darkred}{51.6}/49.7 & \textcolor{darkred}{53.7}/\textcolor{darkred}{52.1} & 45.2/49.8 & 60.0/\textcolor{darkred}{58.3} & \textcolor{darkred}{56.2}/\textcolor{darkred}{54.1} & \textcolor{darkred}{57.3}/\textcolor{darkred}{56.5} & \textcolor{darkred}{52.4}/\textcolor{darkred}{51.2} & 41.2/\textcolor{darkred}{48.7} & 52.4/\textcolor{darkred}{60.3} & 41.7/\textcolor{darkred}{57.4} & 40.6/52.8 & 55.7/\textcolor{darkred}{55.1} & 34.8/48.4 \\ 
\rowcolor{gray!8}Reg2Inv$_{\mathcal{PC}^{2}\text{-}AD}$ & \textcolor{darkred}{89.1}/94.2 & 66.7/91.6 & 88.1/97.4 & 79.5/\textcolor{darkred}{92.1} & 85.7/92.6 & 100.0/99.6 & \textcolor{darkred}{89.3}/\textcolor{darkred}{93.1} & \textcolor{darkred}{93.3}/\textcolor{darkred}{98.1} & 79.7/\textcolor{darkred}{93.5} & 99.5/95.1 & 88.4/\textcolor{darkred}{93.2} & 82.1/\textcolor{darkred}{98.3} & 100.0/99.6 & 100.0/99.4 \\ 
\rowcolor{gray!8}BTF$_{\mathcal{PC}^{2}\text{-}AD}$ & \textcolor{darkred}{44.9}/\textcolor{darkred}{66.4} & 52.3/\textcolor{darkred}{66.8} & 70.5/\textcolor{darkred}{66.6} & 40.0/\textcolor{darkred}{85.3} & 53.8/\textcolor{darkred}{50.0} & \textcolor{darkred}{72.9}/\textcolor{darkred}{82.4} & \textcolor{darkred}{64.9}/\textcolor{darkred}{55.6} & \textcolor{darkred}{73.3}/\textcolor{darkred}{68.8} & 56.5/\textcolor{darkred}{76.8} & 58.6/\textcolor{darkred}{51.2} & \textcolor{darkred}{58.3}/\textcolor{darkred}{79.5} & \textcolor{darkred}{50.6}/\textcolor{darkred}{66.1} & 45.2/\textcolor{darkred}{85.9} & \textcolor{darkred}{65.2}/\textcolor{darkred}{80.6} \\ 
\rowcolor{gray!8}M3DM$_{\mathcal{PC}^{2}\text{-}AD}$ & \textcolor{darkred}{57.5}/\textcolor{darkred}{65.9} & 54.4/\textcolor{darkred}{61.1} & 51.6/56.9 & \textcolor{darkred}{58.1}/\textcolor{darkred}{66.3} & 39.0/\textcolor{darkred}{53.0} & \textcolor{darkred}{58.6}/\textcolor{darkred}{43.4} & \textcolor{darkred}{48.9}/\textcolor{darkred}{53.0} & 47.6/\textcolor{darkred}{56.1} & \textcolor{darkred}{51.3}/\textcolor{darkred}{57.7} & \textcolor{darkred}{54.8}/\textcolor{darkred}{52.0} & 49.9/\textcolor{darkred}{67.9} & 54.2/\textcolor{darkred}{61.0} & \textcolor{darkred}{59.0}/\textcolor{darkred}{68.3} & 55.2/61.5 \\ 
\rowcolor{gray!8}Point-BERT$_{\mathcal{PC}^{2}\text{-}AD}$ & 41.1/57.2 & 50.2/57.5 & \textcolor{darkred}{68.1}/\textcolor{darkred}{58.1} & 50.0/\textcolor{darkred}{64.0} & 51.9/\textcolor{darkred}{59.2} & 52.4/50.6 & \textcolor{darkred}{51.1}/\textcolor{darkred}{55.1} & \textcolor{darkred}{54.8}/\textcolor{darkred}{55.7} & \textcolor{darkred}{68.1}/\textcolor{darkred}{58.4} & 46.2/\textcolor{darkred}{58.6} & 26.7/56.1 & \textcolor{darkred}{52.7}/\textcolor{darkred}{56.9} & 53.3/65.3 & 36.7/59.3 \\ 
\hline 
\hline 
 
\textbf{Method} & \textbf{shelf0} & \textbf{tap0} & \textbf{tap1} & \textbf{vase0} & \textbf{vase1} & \textbf{vase2} & \textbf{vase3} & \textbf{vase4} & \textbf{vase5} & \textbf{vase7} & \textbf{vase8} & \textbf{vase9} & \multicolumn{2}{|c}{\textbf{Average}} \\ 
\hline 
 
PatchCore & \textcolor{baselineblue}{73.7}/78.9 & 40.9/40.5 & \textcolor{baselineblue}{46.9}/54.6 & \textcolor{baselineblue}{78.9}/44.5 & 51.6/29.7 & \textcolor{baselineblue}{71.1}/70.1 & 49.2/52.5 & 57.5/32.7 & \textcolor{baselineblue}{51.6}/35.7 & 50.0/74.9 & \textcolor{baselineblue}{55.2}/44.0 & \textcolor{baselineblue}{62.2}/39.5 & \multicolumn{2}{|c}{62.5/56.6} \\ 
MC3D-AD & \textcolor{baselineblue}{68.4}/58.6 & 29.7/41.8 & \textcolor{baselineblue}{63.7}/\textcolor{baselineblue}{55.6} & \textcolor{baselineblue}{56.7}/\textcolor{baselineblue}{50.9} & \textcolor{baselineblue}{54.8}/52.7 & \textcolor{baselineblue}{52.4}/45.4 & \textcolor{baselineblue}{58.2}/\textcolor{baselineblue}{58.6} & 46.1/50.5 & 56.7/49.5 & \textcolor{baselineblue}{58.6}/\textcolor{baselineblue}{54.7} & \textcolor{baselineblue}{53.0}/\textcolor{baselineblue}{53.3} & 44.8/49.6 & \multicolumn{2}{|c}{52.8/52.4} \\ 
Reg2Inv & 61.4/78.1 & 81.2/\textcolor{baselineblue}{95.6} & \textcolor{baselineblue}{74.4}/\textcolor{baselineblue}{83.5} & \textcolor{baselineblue}{97.5}/\textcolor{baselineblue}{97.7} & \textcolor{baselineblue}{86.7}/\textcolor{baselineblue}{83.0} & 100.0/99.8 & 76.7/90.2 & \textcolor{baselineblue}{75.8}/\textcolor{baselineblue}{95.5} & \textcolor{baselineblue}{100.0}/92.4 & 72.4/92.5 & 85.2/92.7 & 79.7/96.1 & \multicolumn{2}{|c}{86.0/92.4} \\ 
BTF & 53.8/68.5 & 34.2/42.6 & \textcolor{baselineblue}{61.1}/\textcolor{baselineblue}{57.7} & \textcolor{baselineblue}{74.2}/\textcolor{baselineblue}{56.1} & 42.4/38.8 & 40.0/54.3 & 39.7/59.1 & \textcolor{baselineblue}{60.6}/53.1 & 39.0/48.5 & 51.0/55.7 & 25.5/36.6 & 44.2/43.3 & \multicolumn{2}{|c}{53.9/53.8} \\ 
M3DM & \textcolor{baselineblue}{46.1}/\textcolor{baselineblue}{62.0} & \textcolor{baselineblue}{50.6}/\textcolor{baselineblue}{45.3} & \textcolor{baselineblue}{57.4}/44.5 & \textcolor{baselineblue}{57.1}/\textcolor{baselineblue}{64.5} & 64.3/59.4 & 55.2/48.5 & 47.3/59.6 & \textcolor{baselineblue}{66.4}/\textcolor{baselineblue}{42.9} & 44.3/46.0 & \textcolor{baselineblue}{55.2}/36.8 & \textcolor{baselineblue}{65.5}/42.6 & 39.7/47.4 & \multicolumn{2}{|c}{52.0/52.0} \\ 
Point-BERT & 37.4/38.5 & \textcolor{baselineblue}{58.2}/\textcolor{baselineblue}{57.6} & 45.2/\textcolor{baselineblue}{52.5} & \textcolor{baselineblue}{54.6}/57.5 & 43.8/69.3 & 61.0/56.3 & 33.9/55.8 & 39.7/46.9 & \textcolor{baselineblue}{41.9}/\textcolor{baselineblue}{52.5} & 44.8/48.8 & \textcolor{baselineblue}{50.0}/58.9 & \textcolor{baselineblue}{63.0}/50.8 & \multicolumn{2}{|c}{50.3/55.4} \\ 
\rowcolor{gray!8}PatchCore$_{\mathcal{PC}^{2}\text{-}AD}$ & 47.7/\textcolor{darkred}{84.4} & \textcolor{darkred}{41.6}/\textcolor{darkred}{62.6} & 39.8/\textcolor{darkred}{55.8} & 64.8/\textcolor{darkred}{55.8} & \textcolor{darkred}{60.5}/\textcolor{darkred}{76.7} & 68.7/\textcolor{darkred}{93.9} & \textcolor{darkred}{60.2}/\textcolor{darkred}{87.8} & \textcolor{darkred}{57.9}/\textcolor{darkred}{58.8} & 50.9/\textcolor{darkred}{51.6} & \textcolor{darkred}{62.4}/\textcolor{darkred}{94.1} & 51.6/\textcolor{darkred}{96.2} & 59.0/\textcolor{darkred}{55.8} & \multicolumn{2}{|c}{\cellcolor{gray!15}{59.2/77.3 ($\uparrow$ 8.7)}} \\ 
\rowcolor{gray!8}MC3D-AD$_{\mathcal{PC}^{2}\text{-}AD}$ & 59.1/\textcolor{darkred}{60.3} & \textcolor{darkred}{57.9}/\textcolor{darkred}{61.1} & 43.7/44.8 & 45.8/49.3 & 49.5/\textcolor{darkred}{55.4} & 49.0/\textcolor{darkred}{51.5} & 57.9/56.0 & \textcolor{darkred}{58.2}/\textcolor{darkred}{53.1} & \textcolor{darkred}{64.3}/\textcolor{darkred}{59.5} & 56.2/53.9 & 33.3/46.7 & \textcolor{darkred}{57.3}/\textcolor{darkred}{52.8} & \multicolumn{2}{|c}{\cellcolor{gray!15}{52.5/54.3 ($\uparrow$ 0.8)}} \\ 
\rowcolor{gray!8}Reg2Inv$_{\mathcal{PC}^{2}\text{-}AD}$ & \textcolor{darkred}{71.3}/\textcolor{darkred}{84.9} & \textcolor{darkred}{82.7}/91.9 & 66.3/83.1 & 97.1/97.6 & 66.2/76.0 & 100.0/99.8 & \textcolor{darkred}{85.2}/\textcolor{darkred}{90.4} & 72.7/95.4 & 98.1/\textcolor{darkred}{94.1} & \textcolor{darkred}{82.9}/\textcolor{darkred}{96.0} & \textcolor{darkred}{85.8}/\textcolor{darkred}{97.7} & \textcolor{darkred}{87.3}/\textcolor{darkred}{96.3} & \multicolumn{2}{|c}{\cellcolor{gray!15}{87.4/93.1 ($\uparrow$ 1.1)}} \\ 
\rowcolor{gray!8}BTF$_{\mathcal{PC}^{2}\text{-}AD}$ & \textcolor{darkred}{73.9}/\textcolor{darkred}{79.3} & \textcolor{darkred}{43.6}/\textcolor{darkred}{62.4} & 55.6/56.4 & 37.1/55.4 & \textcolor{darkred}{43.8}/\textcolor{darkred}{72.8} & \textcolor{darkred}{57.6}/\textcolor{darkred}{81.4} & \textcolor{darkred}{58.5}/\textcolor{darkred}{82.8} & 52.4/53.1 & \textcolor{darkred}{50.0}/\textcolor{darkred}{56.2} & \textcolor{darkred}{63.8}/\textcolor{darkred}{76.1} & \textcolor{darkred}{33.9}/\textcolor{darkred}{84.6} & \textcolor{darkred}{46.4}/\textcolor{darkred}{58.3} & \multicolumn{2}{|c}{\cellcolor{gray!15}{57.8/71.5 ($\uparrow$ 10.8)}} \\ 
\rowcolor{gray!8}M3DM$_{\mathcal{PC}^{2}\text{-}AD}$ & 38.0/57.2 & 50.0/41.4 & 48.1/\textcolor{darkred}{48.8} & 52.9/61.2 & \textcolor{darkred}{66.2}/\textcolor{darkred}{67.8} & \textcolor{darkred}{60.0}/\textcolor{darkred}{63.5} & \textcolor{darkred}{53.3}/\textcolor{darkred}{62.2} & 58.5/42.0 & \textcolor{darkred}{53.3}/\textcolor{darkred}{54.5} & 49.5/\textcolor{darkred}{41.4} & 51.8/\textcolor{darkred}{65.4} & \textcolor{darkred}{41.5}/\textcolor{darkred}{49.6} & \multicolumn{2}{|c}{\cellcolor{gray!15}{51.1/57.9 ($\uparrow$ 2.5)}} \\ 
\rowcolor{gray!8}Point-BERT$_{\mathcal{PC}^{2}\text{-}AD}$ & \textcolor{darkred}{60.0}/\textcolor{darkred}{55.9} & 45.8/44.3 & \textcolor{darkred}{52.6}/48.8 & 52.9/\textcolor{darkred}{62.7} & \textcolor{darkred}{58.6}/\textcolor{darkred}{73.7} & 61.0/\textcolor{darkred}{58.5} & \textcolor{darkred}{64.5}/\textcolor{darkred}{66.5} & \textcolor{darkred}{63.0}/\textcolor{darkred}{52.9} & 39.5/41.7 & \textcolor{darkred}{54.3}/\textcolor{darkred}{52.4} & 41.5/\textcolor{darkred}{60.0} & 34.2/\textcolor{darkred}{55.3} & \multicolumn{2}{|c}{\cellcolor{gray!15}{51.0/58.0 ($\uparrow$ 1.7)}} \\ 
\hline 
\end{tabular} 
\end{adjustbox} 
 
\end{table*}

\begin{table*}[!t]
\centering

\caption{Per-category O-AUROC/P-AUROC~($\uparrow$) results (\%) on Anomaly-ShapeNet-new. Values in parentheses give the change in Avg.\ AUROC, the mean of O-AUROC and P-AUROC, in percentage points.}
\label{tab:asn_new_pcd_results}

\begin{adjustbox}{max width=\textwidth}
\begin{tabular}{l|ccccccccccccc}
\hline
\multicolumn{14}{c}{\textbf{O-AUROC/P-AUROC}} \\
\hline
\textbf{Method}
& \textbf{cabinet0}
& \textbf{cap1}
& \textbf{cap2}
& \textbf{chair0}
& \textbf{cup2}
& \textbf{desk0}
& \textbf{knife0}
& \textbf{knife1}
& \textbf{microphone1}
& \textbf{screen0}
& \textbf{vase10}
& \textbf{vase6}
& \multicolumn{1}{|c}{\textbf{Average}} \\
\hline

PatchCore
& 54.0/76.8
& \textcolor{baselineblue}{81.6}/92.2
& \textcolor{baselineblue}{73.9}/47.3
& 46.6/73.1
& 69.8/59.2
& 44.6/84.0
& \textcolor{baselineblue}{62.1}/60.3
& 51.3/83.0
& 57.8/48.4
& 47.7/78.8
& \textcolor{baselineblue}{81.9}/76.0
& 53.2/69.5
& \multicolumn{1}{|c}{60.4/70.7} \\

MC3D-AD
& 39.7/56.6
& 53.0/47.6
& 52.3/\textcolor{baselineblue}{46.2}
& \textcolor{baselineblue}{37.7}/50.0
& 50.0/53.0
& \textcolor{baselineblue}{58.0}/\textcolor{baselineblue}{75.7}
& \textcolor{baselineblue}{48.7}/62.0
& \textcolor{baselineblue}{51.0}/55.9
& 45.4/\textcolor{baselineblue}{53.9}
& 40.0/67.1
& \textcolor{baselineblue}{44.1}/\textcolor{baselineblue}{51.3}
& \textcolor{baselineblue}{50.9}/\textcolor{baselineblue}{44.8}
& \multicolumn{1}{|c}{47.6/55.3} \\

Reg2Inv
& 53.7/80.3
& 52.6/80.1
& \textcolor{baselineblue}{74.4}/74.5
& \textcolor{baselineblue}{76.3}/83.8
& \textcolor{baselineblue}{89.5}/91.0
& \textcolor{baselineblue}{59.0}/82.3
& 50.7/61.2
& \textcolor{baselineblue}{65.7}/\textcolor{baselineblue}{76.3}
& 81.7/\textcolor{baselineblue}{99.1}
& 50.4/\textcolor{baselineblue}{82.6}
& 90.7/91.8
& \textcolor{baselineblue}{84.8}/89.6
& \multicolumn{1}{|c}{69.1/82.7} \\

BTF
& 48.3/55.0
& 65.6/78.1
& 47.4/44.6
& 45.3/53.3
& 51.9/60.7
& \textcolor{baselineblue}{54.0}/56.5
& 54.0/61.5
& 51.3/60.0
& 59.2/42.3
& 50.4/59.1
& 55.7/62.1
& 60.3/56.5
& \multicolumn{1}{|c}{53.6/57.5} \\

M3DM
& 50.0/43.2
& \textcolor{baselineblue}{64.2}/62.4
& 23.9/\textcolor{baselineblue}{59.3}
& \textcolor{baselineblue}{47.7}/53.5
& 44.3/37.3
& \textcolor{baselineblue}{48.3}/57.2
& \textcolor{baselineblue}{39.7}/\textcolor{baselineblue}{48.8}
& 46.0/\textcolor{baselineblue}{56.8}
& \textcolor{baselineblue}{48.8}/49.4
& \textcolor{baselineblue}{53.7}/54.0
& \textcolor{baselineblue}{60.0}/39.0
& 53.6/41.9
& \multicolumn{1}{|c}{48.3/50.2} \\

Point-BERT
& 52.3/\textcolor{baselineblue}{52.5}
& \textcolor{baselineblue}{51.6}/60.7
& \textcolor{baselineblue}{54.4}/\textcolor{baselineblue}{67.0}
& \textcolor{baselineblue}{67.3}/45.8
& 41.0/56.6
& \textcolor{baselineblue}{38.0}/\textcolor{baselineblue}{59.1}
& 43.0/\textcolor{baselineblue}{43.1}
& 46.3/42.5
& \textcolor{baselineblue}{58.8}/\textcolor{baselineblue}{65.3}
& 49.3/43.3
& \textcolor{baselineblue}{63.5}/52.8
& \textcolor{baselineblue}{57.6}/51.2
& \multicolumn{1}{|c}{51.9/53.3} \\

\rowcolor{gray!8}
PatchCore$_{\mathcal{PC}^{2}\text{-}AD}$
& \textcolor{darkred}{72.9}/\textcolor{darkred}{93.3}
& 80.9/\textcolor{darkred}{94.8}
& 36.6/\textcolor{darkred}{54.7}
& \textcolor{darkred}{58.4}/\textcolor{darkred}{81.7}
& \textcolor{darkred}{77.3}/\textcolor{darkred}{91.5}
& \textcolor{darkred}{54.6}/\textcolor{darkred}{93.6}
& 45.8/\textcolor{darkred}{81.4}
& \textcolor{darkred}{56.0}/\textcolor{darkred}{90.6}
& \textcolor{darkred}{70.6}/\textcolor{darkred}{98.4}
& \textcolor{darkred}{50.6}/\textcolor{darkred}{90.2}
& 56.8/\textcolor{darkred}{92.9}
& \textcolor{darkred}{59.0}/\textcolor{darkred}{94.6}
& \multicolumn{1}{|c}{
  \cellcolor{gray!15}{60.0/88.2 ($\uparrow$ 8.6)}
} \\

\rowcolor{gray!8}
MC3D-AD$_{\mathcal{PC}^{2}\text{-}AD}$
& \textcolor{darkred}{66.0}/\textcolor{darkred}{72.7}
& \textcolor{darkred}{64.9}/\textcolor{darkred}{55.8}
& \textcolor{darkred}{54.0}/41.5
& 29.3/\textcolor{darkred}{50.6}
& \textcolor{darkred}{58.6}/\textcolor{darkred}{54.9}
& 37.3/71.1
& 40.3/\textcolor{darkred}{63.0}
& 46.3/\textcolor{darkred}{62.4}
& \textcolor{darkred}{46.7}/50.5
& \textcolor{darkred}{56.3}/\textcolor{darkred}{76.4}
& 40.3/45.3
& 41.2/41.7
& \multicolumn{1}{|c}{
  \cellcolor{gray!15}{48.4/57.2 ($\uparrow$ 1.4)}
} \\

\rowcolor{gray!8}
Reg2Inv$_{\mathcal{PC}^{2}\text{-}AD}$
& \textcolor{darkred}{64.7}/\textcolor{darkred}{86.2}
& \textcolor{darkred}{64.2}/\textcolor{darkred}{89.6}
& 72.3/\textcolor{darkred}{74.6}
& 73.7/\textcolor{darkred}{84.6}
& 87.1/\textcolor{darkred}{93.5}
& 58.7/\textcolor{darkred}{83.0}
& \textcolor{darkred}{55.0}/\textcolor{darkred}{63.4}
& 58.3/75.9
& \textcolor{darkred}{84.2}/99.0
& \textcolor{darkred}{68.5}/80.1
& \textcolor{darkred}{93.0}/\textcolor{darkred}{92.0}
& 78.5/\textcolor{darkred}{92.9}
& \multicolumn{1}{|c}{
  \cellcolor{gray!15}{71.5/84.6 ($\uparrow$ 2.2)}
} \\

\rowcolor{gray!8}
BTF$_{\mathcal{PC}^{2}\text{-}AD}$
& \textcolor{darkred}{53.3}/\textcolor{darkred}{61.9}
& \textcolor{darkred}{70.9}/\textcolor{darkred}{89.6}
& \textcolor{darkred}{60.7}/\textcolor{darkred}{53.8}
& \textcolor{darkred}{61.3}/\textcolor{darkred}{66.4}
& \textcolor{darkred}{62.4}/\textcolor{darkred}{84.8}
& 48.0/\textcolor{darkred}{81.0}
& \textcolor{darkred}{54.7}/\textcolor{darkred}{68.6}
& \textcolor{darkred}{54.3}/\textcolor{darkred}{79.2}
& \textcolor{darkred}{62.1}/\textcolor{darkred}{86.7}
& 50.4/\textcolor{darkred}{77.9}
& \textcolor{darkred}{67.0}/\textcolor{darkred}{73.3}
& \textcolor{darkred}{74.5}/\textcolor{darkred}{87.3}
& \multicolumn{1}{|c}{
  \cellcolor{gray!15}{60.0/75.9 ($\uparrow$ 12.4)}
} \\

\rowcolor{gray!8}
M3DM$_{\mathcal{PC}^{2}\text{-}AD}$
& \textcolor{darkred}{55.3}/\textcolor{darkred}{53.2}
& 46.7/\textcolor{darkred}{73.7}
& \textcolor{darkred}{31.6}/56.6
& 47.3/\textcolor{darkred}{60.1}
& \textcolor{darkred}{53.3}/\textcolor{darkred}{50.3}
& 48.0/\textcolor{darkred}{63.1}
& 36.3/46.0
& \textcolor{darkred}{46.3}/46.4
& 48.3/\textcolor{darkred}{59.6}
& 51.1/\textcolor{darkred}{62.0}
& 40.0/\textcolor{darkred}{62.3}
& \textcolor{darkred}{60.6}/\textcolor{darkred}{56.7}
& \multicolumn{1}{|c}{
  \cellcolor{gray!15}{47.1/57.5 ($\uparrow$ 3.1)}
} \\

\rowcolor{gray!8}
Point-BERT$_{\mathcal{PC}^{2}\text{-}AD}$
& \textcolor{darkred}{58.0}/51.4
& 42.1/\textcolor{darkred}{66.5}
& 40.4/59.7
& 64.0/\textcolor{darkred}{58.2}
& \textcolor{darkred}{56.7}/\textcolor{darkred}{58.6}
& 25.3/50.9
& \textcolor{darkred}{50.7}/40.1
& \textcolor{darkred}{53.3}/\textcolor{darkred}{54.4}
& 49.2/61.6
& \textcolor{darkred}{70.7}/\textcolor{darkred}{52.7}
& 53.9/\textcolor{darkred}{63.4}
& 51.8/\textcolor{darkred}{55.8}
& \multicolumn{1}{|c}{
  \cellcolor{gray!15}{51.3/56.1 ($\uparrow$ 1.1)}
} \\

\hline
\end{tabular}
\end{adjustbox}

\end{table*}

\subsubsection{Results on Real3D-AD}

On Real3D-AD, four of the six detectors improve in Avg.\ AUROC (Table~\ref{tab:real3dad_results}): MC3D-AD gains 5.8 percentage points, BTF and M3DM each gain 3.7, and Point-BERT gains 3.2. PatchCore and Reg2Inv decrease by 1.5 and 0.3 percentage points, respectively. PatchCore's O-AUROC increases from 53.0\% to 56.1\%, while its P-AUROC decreases from 71.9\% to 65.8\%, showing that compensation affects detection and localization differently. 

\begin{table*}[!t]
\centering

\caption{Per-category O-AUROC/P-AUROC~($\uparrow$) results (\%) on Real3D-AD. Values in parentheses give the change in Avg.\ AUROC, the mean of O-AUROC and P-AUROC, in percentage points.}
\label{tab:real3dad_results}

\begin{adjustbox}{max width=\textwidth}
\begin{tabular}{l|ccccccccccccc}
\hline
\multicolumn{14}{c}{\textbf{O-AUROC/P-AUROC}} \\
\hline
\textbf{Method}
& \textbf{Airplane}
& \textbf{Car}
& \textbf{Candybar}
& \textbf{Chicken}
& \textbf{Diamond}
& \textbf{Duck}
& \textbf{Fish}
& \textbf{Gemstone}
& \textbf{Seahorse}
& \textbf{Shell}
& \textbf{Starfish}
& \textbf{Toffees}
& \multicolumn{1}{|c}{\textbf{Average}} \\
\hline

PatchCore
& 41.2/\textcolor{baselineblue}{64.2}
& \textcolor{baselineblue}{62.4}/\textcolor{baselineblue}{76.4}
& 51.2/86.3
& 45.4/\textcolor{baselineblue}{60.5}
& 51.1/\textcolor{baselineblue}{90.9}
& 49.3/\textcolor{baselineblue}{66.3}
& 51.2/\textcolor{baselineblue}{79.8}
& \textcolor{baselineblue}{51.5}/\textcolor{baselineblue}{84.6}
& \textcolor{baselineblue}{54.0}/53.7
& \textcolor{baselineblue}{61.5}/\textcolor{baselineblue}{65.7}
& \textcolor{baselineblue}{59.0}/\textcolor{baselineblue}{56.4}
& 58.7/\textcolor{baselineblue}{78.4}
& \multicolumn{1}{|c}{53.0/71.9} \\

MC3D-AD
& 64.6/\textcolor{baselineblue}{66.7}
& 40.8/52.9
& 51.8/49.3
& 53.3/\textcolor{baselineblue}{60.5}
& 64.2/61.4
& \textcolor{baselineblue}{54.0}/50.0
& 58.3/66.1
& \textcolor{baselineblue}{59.7}/\textcolor{baselineblue}{61.1}
& 47.1/50.6
& 44.7/52.1
& 52.4/49.2
& 55.9/68.0
& \multicolumn{1}{|c}{53.9/57.3} \\

Reg2Inv
& \textcolor{baselineblue}{95.3}/96.6
& \textcolor{baselineblue}{84.6}/98.8
& \textcolor{baselineblue}{98.0}/\textcolor{baselineblue}{97.4}
& 95.5/\textcolor{baselineblue}{92.3}
& 100.0/99.1
& 69.5/93.8
& \textcolor{baselineblue}{62.7}/84.5
& \textcolor{baselineblue}{84.2}/98.4
& \textcolor{baselineblue}{57.4}/\textcolor{baselineblue}{61.8}
& 95.7/98.2
& \textcolor{baselineblue}{90.0}/\textcolor{baselineblue}{90.3}
& 75.2/89.8
& \multicolumn{1}{|c}{84.0/91.8} \\

BTF
& 44.7/70.6
& \textcolor{baselineblue}{53.6}/69.4
& 48.5/70.4
& \textcolor{baselineblue}{51.9}/\textcolor{baselineblue}{75.1}
& \textcolor{baselineblue}{68.7}/84.5
& 58.0/82.8
& \textcolor{baselineblue}{56.3}/69.4
& \textcolor{baselineblue}{68.4}/82.6
& 45.7/51.0
& 51.2/55.1
& 49.9/\textcolor{baselineblue}{55.0}
& 58.7/81.5
& \multicolumn{1}{|c}{54.6/70.6} \\

M3DM
& 31.7/49.2
& 47.0/50.5
& 42.2/60.8
& 49.9/65.0
& \textcolor{baselineblue}{65.7}/\textcolor{baselineblue}{60.3}
& 47.0/61.8
& \textcolor{baselineblue}{60.6}/59.4
& \textcolor{baselineblue}{66.0}/65.0
& 50.9/56.7
& 62.9/\textcolor{baselineblue}{73.7}
& \textcolor{baselineblue}{46.6}/59.2
& 51.1/64.7
& \multicolumn{1}{|c}{51.8/60.5} \\

Point-BERT
& 29.0/\textcolor{baselineblue}{59.7}
& 50.6/45.2
& 52.2/56.1
& 47.8/\textcolor{baselineblue}{64.4}
& 49.4/53.0
& \textcolor{baselineblue}{54.0}/43.6
& \textcolor{baselineblue}{70.4}/64.1
& 50.0/\textcolor{baselineblue}{55.7}
& 58.2/59.0
& \textcolor{baselineblue}{52.5}/73.5
& \textcolor{baselineblue}{64.4}/55.7
& 55.5/69.3
& \multicolumn{1}{|c}{52.8/58.3} \\

\rowcolor{gray!8}
PatchCore$_{\mathcal{PC}^{2}\text{-}AD}$
& \textcolor{darkred}{51.2}/62.1
& 59.7/71.9
& \textcolor{darkred}{66.5}/\textcolor{darkred}{88.3}
& \textcolor{darkred}{54.2}/50.2
& \textcolor{darkred}{63.0}/79.8
& \textcolor{darkred}{54.8}/49.2
& \textcolor{darkred}{64.6}/72.0
& 42.9/67.9
& 52.5/\textcolor{darkred}{56.0}
& 59.4/64.3
& 42.0/55.7
& \textcolor{darkred}{61.9}/72.2
& \multicolumn{1}{|c}{
  \cellcolor{gray!15}{56.1/65.8 ($\downarrow$ 1.5)}
} \\

\rowcolor{gray!8}
MC3D-AD$_{\mathcal{PC}^{2}\text{-}AD}$
& \textcolor{darkred}{72.2}/62.9
& \textcolor{darkred}{48.3}/\textcolor{darkred}{68.4}
& \textcolor{darkred}{58.1}/\textcolor{darkred}{62.4}
& \textcolor{darkred}{64.0}/58.3
& \textcolor{darkred}{70.9}/\textcolor{darkred}{68.8}
& 52.2/\textcolor{darkred}{55.3}
& \textcolor{darkred}{68.0}/\textcolor{darkred}{76.5}
& 45.5/38.5
& \textcolor{darkred}{69.7}/\textcolor{darkred}{56.8}
& \textcolor{darkred}{59.2}/\textcolor{darkred}{59.1}
& \textcolor{darkred}{52.5}/\textcolor{darkred}{58.4}
& \textcolor{darkred}{73.7}/\textcolor{darkred}{74.3}
& \multicolumn{1}{|c}{
  \cellcolor{gray!15}{61.2/61.6 ($\uparrow$ 5.8)}
} \\

\rowcolor{gray!8}
Reg2Inv$_{\mathcal{PC}^{2}\text{-}AD}$
& 93.8/96.6
& 84.0/\textcolor{darkred}{99.0}
& 96.1/96.7
& \textcolor{darkred}{95.9}/92.2
& 100.0/99.1
& \textcolor{darkred}{73.6}/\textcolor{darkred}{94.4}
& 61.4/\textcolor{darkred}{84.6}
& 81.5/98.4
& 55.5/61.3
& \textcolor{darkred}{96.2}/\textcolor{darkred}{98.4}
& 88.9/88.8
& \textcolor{darkred}{76.6}/\textcolor{darkred}{90.5}
& \multicolumn{1}{|c}{
  \cellcolor{gray!15}{83.6/91.7 ($\downarrow$ 0.3)}
} \\

\rowcolor{gray!8}
BTF$_{\mathcal{PC}^{2}\text{-}AD}$
& \textcolor{darkred}{68.4}/70.6
& 53.2/\textcolor{darkred}{75.4}
& \textcolor{darkred}{63.8}/\textcolor{darkred}{85.7}
& 46.9/67.5
& 67.6/\textcolor{darkred}{85.7}
& \textcolor{darkred}{58.9}/\textcolor{darkred}{83.2}
& 55.9/\textcolor{darkred}{71.4}
& 62.6/\textcolor{darkred}{89.8}
& \textcolor{darkred}{51.4}/\textcolor{darkred}{53.3}
& \textcolor{darkred}{59.0}/\textcolor{darkred}{60.4}
& \textcolor{darkred}{60.3}/53.4
& \textcolor{darkred}{59.7}/\textcolor{darkred}{85.4}
& \multicolumn{1}{|c}{
  \cellcolor{gray!15}{59.0/73.5 ($\uparrow$ 3.7)}
} \\

\rowcolor{gray!8}
M3DM$_{\mathcal{PC}^{2}\text{-}AD}$
& \textcolor{darkred}{40.5}/\textcolor{darkred}{61.6}
& \textcolor{darkred}{55.9}/\textcolor{darkred}{55.4}
& \textcolor{darkred}{43.7}/\textcolor{darkred}{71.6}
& \textcolor{darkred}{52.9}/\textcolor{darkred}{65.6}
& 61.5/60.2
& \textcolor{darkred}{47.1}/\textcolor{darkred}{65.8}
& 57.2/\textcolor{darkred}{62.1}
& 64.5/\textcolor{darkred}{65.4}
& \textcolor{darkred}{64.4}/\textcolor{darkred}{61.6}
& \textcolor{darkred}{65.0}/73.1
& 46.0/\textcolor{darkred}{62.4}
& \textcolor{darkred}{64.6}/\textcolor{darkred}{66.6}
& \multicolumn{1}{|c}{
  \cellcolor{gray!15}{55.3/64.3 ($\uparrow$ 3.7)}
} \\

\rowcolor{gray!8}
Point-BERT$_{\mathcal{PC}^{2}\text{-}AD}$
& \textcolor{darkred}{41.5}/57.9
& \textcolor{darkred}{51.3}/\textcolor{darkred}{53.2}
& \textcolor{darkred}{56.3}/\textcolor{darkred}{66.2}
& \textcolor{darkred}{53.7}/62.6
& \textcolor{darkred}{69.4}/\textcolor{darkred}{59.7}
& 51.9/\textcolor{darkred}{50.1}
& 64.5/\textcolor{darkred}{70.0}
& \textcolor{darkred}{52.3}/51.7
& \textcolor{darkred}{58.4}/\textcolor{darkred}{65.0}
& 50.8/\textcolor{darkred}{74.8}
& 52.7/\textcolor{darkred}{60.1}
& \textcolor{darkred}{62.2}/\textcolor{darkred}{74.4}
& \multicolumn{1}{|c}{
  \cellcolor{gray!15}{55.4/62.1 ($\uparrow$ 3.2)}
} \\

\hline
\end{tabular}
\end{adjustbox}

\end{table*}

\subsection{Qualitative Results}

\begin{figure*}[!p]
    \centering

    \includegraphics[width=\textwidth]{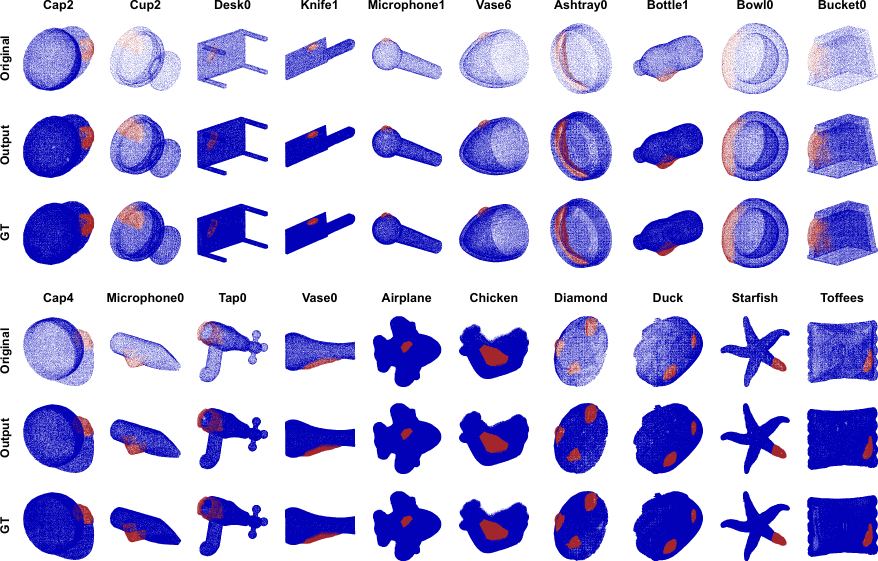}
    \caption{Qualitative results of $\mathcal{PC}^{2}$-AD on Anomaly-ShapeNet and Real3D-AD.}
    \label{fig:qualitative_results}

    \vspace{0.5em}

    \includegraphics[width=\textwidth]{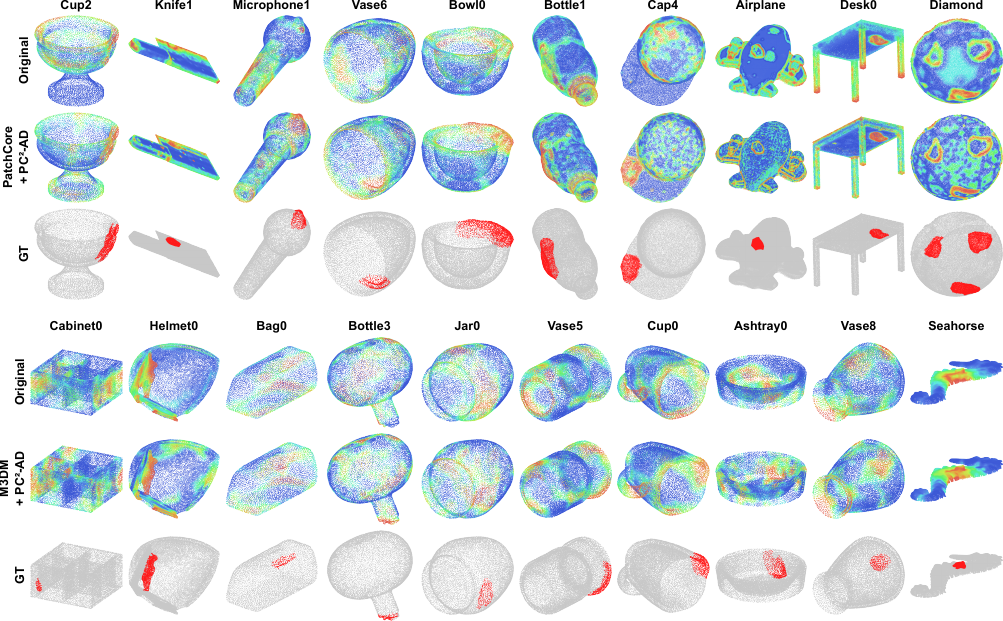}
    \caption{Qualitative anomaly localization results of PatchCore and M3DM before and after applying $\mathcal{PC}^{2}$-AD.}
    \label{fig:qualitative_detector_results}

\end{figure*}

Fig.~\ref{fig:qualitative_results} shows representative examples from Anomaly-ShapeNet and Real3D-AD. From top to bottom, the rows labeled Original, Output, and GT denote the low-resolution input, resolution-compensated output, and original dense point-level ground truth, respectively. Blue and red indicate normal and anomalous regions. For the low-resolution input, point labels are obtained by downsampling the original ground-truth labels using the same FPS indices as the input point cloud. In the resolution-compensated output, the original input points retain these labels, while labels of the added compensation points are transferred from the dense ground truth by nearest neighbor for visualization only. Compared with the low-resolution input, the compensated output provides denser and more continuous surface coverage while leaving the original input points unchanged. The displayed examples suggest that the added coverage remains consistent with the observed local geometry. Additional qualitative examples are provided in Fig.~S2 of the supplementary material.

Fig.~\ref{fig:qualitative_detector_results} presents anomaly localization results of PatchCore and M3DM before and after applying $\mathcal{PC}^{2}$-AD. For each detector, the rows labeled Original, $+\mathcal{PC}^{2}$-AD, and GT denote the anomaly maps obtained from the low-resolution input, resolution-compensated input, and point-level ground truth, respectively. Only the original input points are displayed, while the added compensation points participate in detector inference but are not visualized. Compared with the original low-resolution inputs, the resolution-compensated inputs produce anomaly responses that are more consistent with the corresponding point-level ground truth in the displayed examples. These results suggest that point compensation can improve downstream localization for detectors with different feature representations. Additional qualitative anomaly localization results for Point-BERT, BTF, MC3D-AD, and Reg2Inv are provided in Fig.~S3 of the supplementary material.

\afterpage{\clearpage}

\subsection{Ablation Study}

Table~\ref{tab:ablation} reports the component ablation results of $\mathcal{PC}^{2}$-AD on Anomaly-ShapeNet using PatchCore. Starting from the complete model, we sequentially remove NPPC same-anchor replacement, GACF, and TDA. Removing replacement decreases Object-Avg from 62.7\% to 61.8\%, while Point-Avg remains at 44.9\%, suggesting that its clearest contribution in this setting is to object-level performance. Further removing GACF decreases Point-Avg from 44.9\% to 44.1\%, while Object-Avg increases to 62.5\%. This comparison suggests that GACF helps maintain point-level performance along this removal path, although its effect on Object-Avg is opposite. When TDA is also removed, Object-Avg and Point-Avg decrease to 61.7\% and 44.0\%, respectively. This result is consistent with TDA's intended role in adapting candidate generation to the normal geometry of the target domain. Overall, the complete model achieves the highest Object-Avg and joint-highest Point-Avg. Compared with the configuration without TDA, GACF, and NPPC replacement, it improves the two metrics by 1.0 and 0.9 percentage points, respectively. The intermediate results show different component effects on the two metrics along the tested removal path.

\begin{table}[!htbp]
\centering

\caption{Component ablation results (\%) of $\mathcal{PC}^{2}$-AD on
Anomaly-ShapeNet using PatchCore. Object-Avg and Point-Avg
denote the averages of O-AUROC/O-AUPRC and P-AUROC/P-AUPRC, respectively.}
\label{tab:ablation}

\begin{adjustbox}{max width=\linewidth}
\begin{tabular}{lcc}
\toprule
\textbf{Configuration}
& \textbf{Object-Avg}
& \textbf{Point-Avg} \\
\midrule

\textbf{$\mathcal{PC}^{2}$-AD}
& \textbf{62.7}
& \textbf{44.9} \\

$\mathcal{PC}^{2}$-AD w/o NPPC replacement
& 61.8
& \textbf{44.9} \\

$\mathcal{PC}^{2}$-AD w/o GACF \& NPPC replacement
& 62.5
& 44.1 \\

$\mathcal{PC}^{2}$-AD w/o TDA, GACF \& NPPC replacement
& 61.7
& 44.0 \\
\bottomrule
\end{tabular}
\end{adjustbox}

\end{table}

\begin{figure*}[!b]
    \centering
    \includegraphics[width=\textwidth]
    {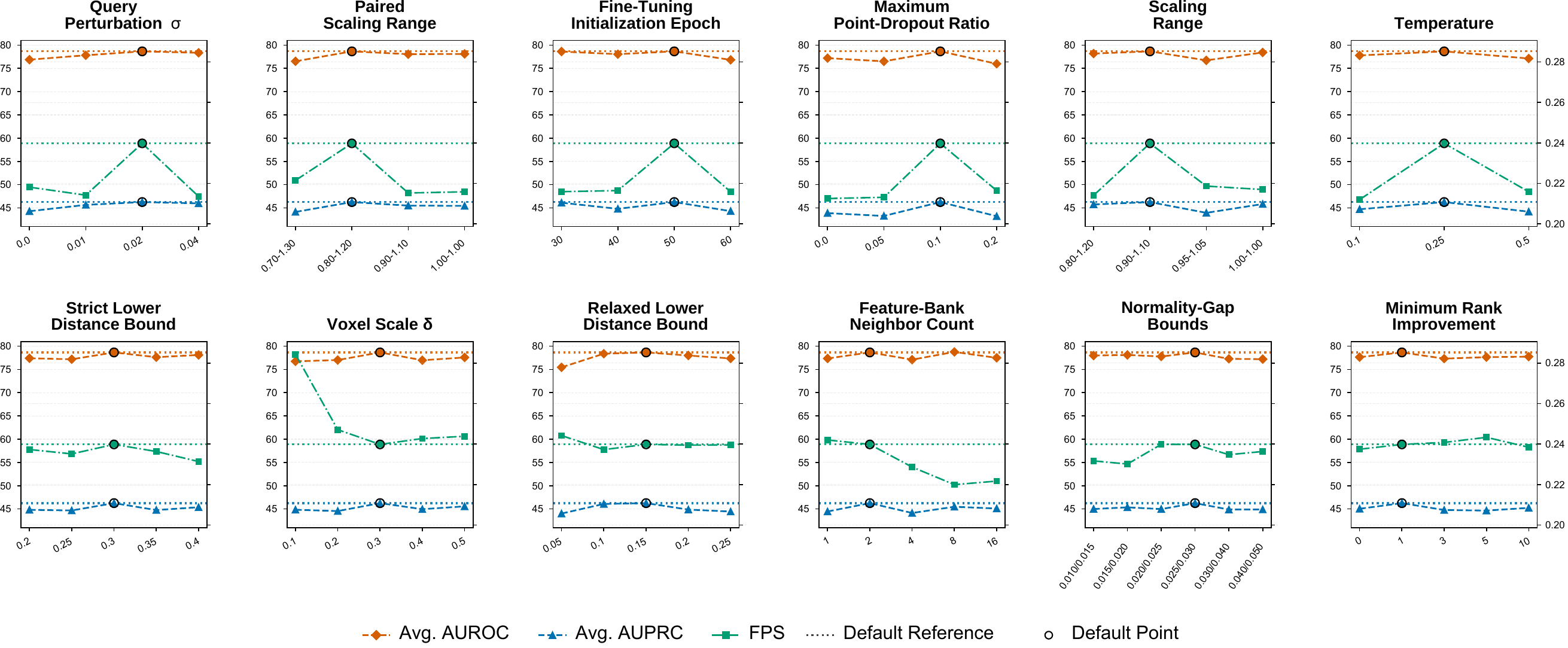}
    \caption{Parameter sensitivity of $\mathcal{PC}^{2}$-AD. Avg.\ AUROC and Avg.\ AUPRC denote the averages of O-AUROC and P-AUROC, and of O-AUPRC and P-AUPRC, respectively. FPS denotes generation speed in samples per second. Hollow black circles mark the default configuration, and dotted horizontal lines indicate the corresponding default results.}
    \label{fig:parameter_sensitivity}
\end{figure*}

\subsection{Parameter Sensitivity}

We conduct parameter sensitivity experiments on the 12-category Anomaly-ShapeNet-new setting using Reg2Inv as the downstream detector. One parameter setting is varied at a time, with jointly constrained parameter pairs varied together, while all others are fixed at their default values. Fig.~\ref{fig:parameter_sensitivity} reports Avg.\ AUROC, Avg.\ AUPRC, and FPS for 12 core parameters, with the remaining 17 parameters provided in Fig.~S1 of the supplementary material. Under the default configuration, single-GPU generation runs at approximately 0.24 samples/s with a peak GPU memory usage of approximately 2.1\,GB. Since memory usage varies only slightly across configurations, the following analysis focuses on detection performance and generation speed.

For TDA, a \textbf{Query Perturbation $\sigma$} of 0.02 and a \textbf{Paired Scaling Range} of $0.80$--$1.20$ achieve the best observed aggregate detection performance among the tested settings. Smaller perturbations or narrower scaling ranges may provide limited variation for adaptation, whereas stronger perturbations or wider ranges tend to degrade detection performance in these sweeps. For NPPC training, the epoch-50 \textbf{Fine-Tuning Initialization Epoch} of CNP performs better than the epoch-60 checkpoint for initializing MSCF. The preferred \textbf{Maximum Point-Dropout Ratio} in CNP is 0.10, while the preferred \textbf{Scaling Range} and \textbf{Temperature $\tau$} in MSCF are $0.90$--$1.10$ and 0.25, respectively. The augmentation results suggest that moderate variation can be more suitable than either weak or excessive perturbation in these training stages.

For GACF, $\alpha_{\mathrm{str}}=0.30$, $\delta=0.30$, and $\alpha_{\mathrm{rel}}=0.15$ achieve the best observed aggregate detection performance among the tested settings. These results are consistent with the intended balance between retaining useful candidates and suppressing spatial redundancy. For same-anchor replacement within NPPC, $K_{\mathrm{nn}}=2$ provides a favorable balance between detection performance and efficiency, with higher generation speed than the larger tested neighbor counts. The normality-gap bounds $(\epsilon_1,\epsilon_2)=(0.025,0.030)$ and minimum rank improvement $\kappa=1$ achieve the best observed aggregate detection performance among the tested settings for normality-guided replacement.

Overall, intermediate values of the geometric and normality parameters generally provide a favorable balance between detection performance and efficiency among the tested settings. The curves also show that detection performance and generation speed can respond differently to a parameter change. These sweeps describe sensitivity around the shared default configuration used in the main experiments.

\section{Conclusion}

We presented $\mathcal{PC}^{2}$-AD to address the train--test sampling-resolution gap motivated by low-resolution sensing in edge deployments. The framework uses normal training geometry to generate dense candidates, filters them by spacing and coverage, and selects compensation points according to local normality consistency with input anchors. The resulting point clouds retain the original observations and can be used by existing downstream detectors. Across six detectors, the mean of object-level and point-level AUROC improves for all six in both Anomaly-ShapeNet settings and for four on Real3D-AD. The results show that point cloud compensation can improve anomaly detection on sparse inputs, with benefits that vary between object-level detection and point-level localization.

\textbf{Limitations:}
The low-resolution inputs in our experiments are constructed through controlled farthest point sampling and therefore do not fully reproduce sensor-specific noise, occlusion, or nonuniform sampling. In addition, the effectiveness of resolution compensation depends on the sensitivity of downstream detectors to changes in sampling density. Future work will investigate more realistic sensor degradation settings and more efficient resolution compensation for practical deployment.

\bibliographystyle{IEEEtran}
\bibliography{refs}

@inproceedings{qi2017pointnet,
  author      = {Qi, Charles R. and Su, Hao and Mo, Kaichun and Guibas, Leonidas J.},
  title       = {{PointNet}: Deep learning on point sets for {3D} classification and segmentation},
  booktitle   = {Proceedings of the IEEE Conference on Computer Vision and Pattern Recognition (CVPR)},
  year        = {2017},
  pages       = {652--660},
  url         = {https://openaccess.thecvf.com/content_cvpr_2017/html/Qi_PointNet_Deep_Learning_CVPR_2017_paper.html}
}

@inproceedings{liang2025taming,
  author      = {Liang, Hanzhe and Zhang, Jie and Dai, Tao and Shen, Linlin and Wang, Jinbao and Gao, Can},
  title       = {Taming anomalies with down-up sampling networks: Group center preserving reconstruction for {3D} anomaly detection},
  booktitle   = {Proceedings of the 33rd ACM International Conference on Multimedia (ACM MM)},
  year        = {2025},
  pages       = {7133--7141},
  doi         = {10.1145/3746027.3754492},
  url         = {https://doi.org/10.1145/3746027.3754492}
}

@inproceedings{wang2023cvpr-multimodal,
  author      = {Wang, Yue and Peng, Jinlong and Zhang, Jiangning and Yi, Ran and Wang, Yabiao and Wang, Chengjie},
  title       = {Multimodal Industrial Anomaly Detection via Hybrid Fusion},
  booktitle   = {Proceedings of the IEEE/CVF Conference on Computer Vision and Pattern Recognition (CVPR)},
  year        = {2023},
  pages       = {8032--8041},
  doi         = {10.1109/CVPR52729.2023.00776},
  url         = {https://openaccess.thecvf.com/content/CVPR2023/html/Wang_Multimodal_Industrial_Anomaly_Detection_via_Hybrid_Fusion_CVPR_2023_paper.html}
}

@inproceedings{ijcai2025p182,
  author      = {Liu, Yi and Zhang, Changsheng and Yang, Yufei},
  title       = {{Template3D-AD}: Point Cloud Template Matching Method Based on Center Points for {3D} Anomaly Detection},
  booktitle   = {Proceedings of the Thirty-Fourth International Joint Conference on Artificial Intelligence (IJCAI)},
  publisher   = {International Joint Conferences on Artificial Intelligence Organization},
  year        = {2025},
  pages       = {1630--1638},
  doi         = {10.24963/ijcai.2025/182},
  url         = {https://www.ijcai.org/proceedings/2025/182}
}

@inproceedings{Yu_2018_CVPR,
  author      = {Yu, Lequan and Li, Xianzhi and Fu, Chi-Wing and Cohen-Or, Daniel and Heng, Pheng-Ann},
  title       = {{PU-Net}: Point Cloud Upsampling Network},
  booktitle   = {Proceedings of the IEEE Conference on Computer Vision and Pattern Recognition (CVPR)},
  year        = {2018},
  pages       = {2790--2799},
  url         = {https://openaccess.thecvf.com/content_cvpr_2018/html/Yu_PU-Net_Point_Cloud_CVPR_2018_paper.html}
}

@inproceedings{Qian_2021_CVPR,
  author      = {Qian, Guocheng and Abualshour, Abdulellah and Li, Guohao and Thabet, Ali and Ghanem, Bernard},
  title       = {{PU-GCN}: Point Cloud Upsampling Using Graph Convolutional Networks},
  booktitle   = {Proceedings of the IEEE/CVF Conference on Computer Vision and Pattern Recognition (CVPR)},
  year        = {2021},
  pages       = {11683--11692},
  url         = {https://openaccess.thecvf.com/content/CVPR2021/html/Qian_PU-GCN_Point_Cloud_Upsampling_Using_Graph_Convolutional_Networks_CVPR_2021_paper.html}
}

@inproceedings{Qiu_2022_ACCV,
  author      = {Qiu, Shi and Anwar, Saeed and Barnes, Nick},
  title       = {{PU-Transformer}: Point Cloud Upsampling Transformer},
  booktitle   = {Proceedings of the Asian Conference on Computer Vision (ACCV)},
  year        = {2022},
  pages       = {2475--2493},
  url         = {https://openaccess.thecvf.com/content/ACCV2022/html/Qiu_PU-Transformer_Point_Cloud_Upsampling_Transformer_ACCV_2022_paper.html}
}

@inproceedings{He_2023_CVPR,
  author      = {He, Yun and Tang, Danhang and Zhang, Yinda and Xue, Xiangyang and Fu, Yanwei},
  title       = {{Grad-PU}: Arbitrary-Scale Point Cloud Upsampling via Gradient Descent With Learned Distance Functions},
  booktitle   = {Proceedings of the IEEE/CVF Conference on Computer Vision and Pattern Recognition (CVPR)},
  year        = {2023},
  pages       = {5354--5363},
  url         = {https://openaccess.thecvf.com/content/CVPR2023/html/He_Grad-PU_Arbitrary-Scale_Point_Cloud_Upsampling_via_Gradient_Descent_With_Learned_CVPR_2023_paper.html}
}

@article{11644920,
  author      = {Liang, Hanzhe and Gao, Can and Niu, Zehai and Zhang, Jie and Wang, Jinbao and Shen, Linlin},
  title       = {An Information-aware Reconstruction Model for Multi-category {3D} Anomaly Detection},
  journal     = {IEEE Transactions on Artificial Intelligence (TAI)},
  year        = {2026},
  pages       = {1--10},
  doi         = {10.1109/TAI.2026.3717889},
  url         = {https://doi.org/10.1109/TAI.2026.3717889}
}

@inproceedings{3746027.3755261,
  author      = {Xiang, An and Huang, Zixuan and Gao, Xitong and Ye, Kejiang and Xu, Cheng-zhong},
  title       = {{BridgeNet}: A Unified Multimodal Framework for Bridging {2D} and {3D} Industrial Anomaly Detection},
  booktitle   = {Proceedings of the 33rd ACM International Conference on Multimedia (ACM MM)},
  year        = {2025},
  pages       = {1579--1587},
  doi         = {10.1145/3746027.3755261},
  url         = {https://doi.org/10.1145/3746027.3755261}
}

@inproceedings{1623272,
  author      = {Bolles, Robert C. and Fischler, Martin A.},
  title       = {A {RANSAC}-based approach to model fitting and its application to finding cylinders in range data},
  booktitle   = {Proceedings of the 7th International Joint Conference on Artificial Intelligence (IJCAI)},
  volume      = {2},
  year        = {1981},
  pages       = {637--643},
  url         = {https://dl.acm.org/doi/10.5555/1623264.1623272}
}

@misc{balapour2026anomalyfactory3dmodular,
  author      = {Ali Balapour and Faraz Hach},
  title       = {Anomaly Factory {3D}: A Modular Framework for Diverse Pseudo-Anomaly Synthesis in Unsupervised {3D} Anomaly Detection},
  year        = {2026},
  doi = {10.48550/arXiv.2606.29181},
  howpublished= {arXiv preprint arXiv:2606.29181},
  url         = {https://arxiv.org/abs/2606.29181}
}

@inproceedings{664647.3680919,
  author      = {Zhu, Hongze and Xie, Guoyang and Hou, Chengbin and Dai, Tao and Gao, Can and Wang, Jinbao and Shen, Linlin},
  title       = {Towards High-resolution {3D} Anomaly Detection via Group-Level Feature Contrastive Learning},
  booktitle   = {Proceedings of the 32nd ACM International Conference on Multimedia (ACM MM)},
  year        = {2024},
  pages       = {4680--4689},
  doi         = {10.1145/3664647.3680919},
  url         = {https://doi.org/10.1145/3664647.3680919}
}

@inproceedings{zha2026casl,
  author      = {Zha, Yaohua and Yuerong, Xue and Fan, Chunlin and Wang, Yuansong and Dai, Tao and Chen, Ke and Xia, Shu-Tao},
  title       = {{CASL}: Curvature-augmented self-supervised learning for {3D} anomaly detection},
  booktitle   = {Proceedings of the AAAI Conference on Artificial Intelligence (AAAI)},
  volume      = {40},
  number      = {15},
  year        = {2026},
  pages       = {12340--12348},
  doi         = {10.1609/aaai.v40i15.38226},
  url         = {https://ojs.aaai.org/index.php/AAAI/article/view/38226}
}

@inproceedings{PO3AD,
  author      = {Ye, Jianan and Zhao, Weiguang and Yang, Xi and Cheng, Guangliang and Huang, Kaizhu},
  title       = {{PO3AD}: Predicting Point Offsets toward Better {3D} Point Cloud Anomaly Detection},
  booktitle   = {Proceedings of the IEEE/CVF Conference on Computer Vision and Pattern Recognition (CVPR)},
  year        = {2025},
  pages       = {1353--1362},
  url         = {https://openaccess.thecvf.com/content/CVPR2025/html/Ye_PO3AD_Predicting_Point_Offsets_toward_Better_3D_Point_Cloud_Anomaly_CVPR_2025_paper.html}
}

@article{GLFM,
  author      = {Cheng, Yuqi and Cao, Yunkang and Wang, Dongfang and Shen, Weiming and Li, Wenlong},
  title       = {Boosting Global-Local Feature Matching via Anomaly Synthesis for Multi-Class Point Cloud Anomaly Detection},
  journal     = {IEEE Transactions on Automation Science and Engineering (TASE)},
  volume      = {22},
  year        = {2025},
  pages       = {12560--12571},
  doi         = {10.1109/TASE.2025.3544462},
  url         = {https://doi.org/10.1109/TASE.2025.3544462}
}

@article{cao2024complementary,
  author      = {Cao, Yunkang and Xu, Xiaohao and Shen, Weiming},
  title       = {Complementary pseudo multimodal feature for point cloud anomaly detection},
  journal     = {Pattern Recognition (PR)},
  volume      = {156},
  publisher   = {Elsevier},
  year        = {2024},
  pages       = {110761},
  doi         = {10.1016/j.patcog.2024.110761},
  url         = {https://doi.org/10.1016/j.patcog.2024.110761}
}

@inproceedings{liang2025look,
  author      = {Liang, Hanzhe and Xie, Guoyang and Hou, Chengbin and Wang, Bingshu and Gao, Can and Wang, Jinbao},
  title       = {Look inside for more: Internal spatial modality perception for {3D} anomaly detection},
  booktitle   = {Proceedings of the AAAI Conference on Artificial Intelligence (AAAI)},
  volume      = {39},
  number      = {5},
  year        = {2025},
  pages       = {5146--5154},
  doi         = {10.1609/aaai.v39i5.32546},
  url         = {https://ojs.aaai.org/index.php/AAAI/article/view/32546}
}

@inproceedings{pang2022masked,
  author      = {Pang, Yatian and Wang, Wenxiao and Tay, Francis E. H. and Liu, Wei and Tian, Yonghong and Yuan, Li},
  title       = {Masked autoencoders for point cloud self-supervised learning},
  booktitle   = {Proceedings of the 17th European Conference on Computer Vision (ECCV), Part II},
  series      = {Lecture Notes in Computer Science},
  volume      = {13662},
  publisher   = {Springer},
  year        = {2022},
  pages       = {604--621},
  doi         = {10.1007/978-3-031-20086-1_35},
  url         = {https://doi.org/10.1007/978-3-031-20086-1_35}
}

@inproceedings{Long_2026_CVPR,
  author      = {Long, Kaifang and Ma, Lianbo and Liu, Jiaqi and Liu, Liming and Xie, Guoyang},
  title       = {Towards an Incremental Unified Multimodal Anomaly Detection: Augmenting Multimodal Denoising From an Information Bottleneck Perspective},
  booktitle   = {Proceedings of the IEEE/CVF Conference on Computer Vision and Pattern Recognition (CVPR)},
  year        = {2026},
  pages       = {14116--14125},
  url         = {https://openaccess.thecvf.com/content/CVPR2026/html/Long_Towards_an_Incremental_Unified_Multimodal_Anomaly_Detection_Augmenting_Multimodal_Denoising_CVPR_2026_paper.html}
}

@inproceedings{11445823,
  author      = {Zheng, Bozhong and Gan, Jinye and Xu, Xiaohao and Chen, Xintao and Li, Wenqiao and Huang, Xiaonan and Ni, Na and Wu, Yingna},
  title       = {Bridging {3D} Anomaly Localization and Repair Via High-Quality Continuous Geometric Representation},
  booktitle   = {Proceedings of the IEEE/CVF International Conference on Computer Vision (ICCV)},
  year        = {2025},
  pages       = {27063--27072},
  doi         = {10.1109/ICCV51701.2025.02512},
  url         = {https://doi.org/10.1109/ICCV51701.2025.02512}
}

@inproceedings{cheng2026towards,
  author      = {Cheng, Yuqi and Sun, Yihan and Zhang, Hui and Shen, Weiming and Cao, Yunkang},
  title       = {Towards high-resolution {3D} anomaly detection: A scalable dataset and real-time framework for subtle industrial defects},
  booktitle   = {Proceedings of the AAAI Conference on Artificial Intelligence (AAAI)},
  volume      = {40},
  number      = {5},
  year        = {2026},
  pages       = {3327--3334},
  doi         = {10.1609/aaai.v40i5.37328},
  url         = {https://ojs.aaai.org/index.php/AAAI/article/view/37328}
}

@inproceedings{anomaly_shapenet,
  author      = {Li, Wenqiao and Xu, Xiaohao and Gu, Yao and Zheng, Bozhong and Gao, Shenghua and Wu, Yingna},
  title       = {Towards Scalable {3D} Anomaly Detection and Localization: A Benchmark via {3D} Anomaly Synthesis and a Self-Supervised Learning Network},
  booktitle   = {Proceedings of the IEEE/CVF Conference on Computer Vision and Pattern Recognition (CVPR)},
  year        = {2024},
  pages       = {22207--22216},
  url         = {https://openaccess.thecvf.com/content/CVPR2024/html/Li_Towards_Scalable_3D_Anomaly_Detection_and_Localization_A_Benchmark_via_CVPR_2024_paper.html}
}

@inproceedings{liu2023neurips-real3dad,
  author      = {Liu, Jiaqi and Xie, Guoyang and Chen, Ruitao and Li, Xinpeng and Wang, Jinbao and Liu, Yong and Wang, Chengjie and Zheng, Feng},
  title       = {{Real3D-AD}: A Dataset of Point Cloud Anomaly Detection},
  booktitle   = {Advances in Neural Information Processing Systems (NeurIPS)},
  volume      = {36},
  year        = {2023},
  pages       = {30402--30415},
  doi         = {10.52202/075280-1324},
  url         = {https://proceedings.neurips.cc/paper_files/paper/2023/hash/611b896d447df43c898062358df4c114-Abstract-Datasets_and_Benchmarks.html}
}

@inproceedings{roth2022patchcore,
  author      = {Roth, Karsten and Pemula, Latha and Zepeda, Joaquin and Sch{\"o}lkopf, Bernhard and Brox, Thomas and Gehler, Peter},
  title       = {Towards Total Recall in Industrial Anomaly Detection},
  booktitle   = {Proceedings of the IEEE/CVF Conference on Computer Vision and Pattern Recognition (CVPR)},
  year        = {2022},
  pages       = {14318--14328},
  url         = {https://openaccess.thecvf.com/content/CVPR2022/html/Roth_Towards_Total_Recall_in_Industrial_Anomaly_Detection_CVPR_2022_paper.html}
}

@inproceedings{cheng2025mc3dad,
  author      = {Cheng, Jiayi and Gao, Can and Zhou, Jie and Wen, Jiajun and Dai, Tao and Wang, Jinbao},
  title       = {{MC3D-AD}: A Unified Geometry-Aware Reconstruction Model for Multi-Category {3D} Anomaly Detection},
  booktitle   = {Proceedings of the Thirty-Fourth International Joint Conference on Artificial Intelligence (IJCAI)},
  year        = {2025},
  pages       = {837--845},
  doi         = {10.24963/ijcai.2025/94},
  url         = {https://www.ijcai.org/proceedings/2025/94}
}

@inproceedings{yu2025reg2inv,
  author      = {Yu, Yuyang and Chen, Zhengwei and Xu, Xuemiao and Zhang, Lei and Yang, Haoxin and Nie, Yongwei and He, Shengfeng},
  title       = {Registration is a Powerful Rotation-Invariance Learner for {3D} Anomaly Detection},
  booktitle   = {Advances in Neural Information Processing Systems (NeurIPS)},
  volume      = {38},
  year        = {2025},
  pages       = {5073--5099},
  doi         = {10.52202/085713-0180},
  url         = {https://proceedings.neurips.cc/paper_files/paper/2025/hash/07bc722f08f096e6ea7ee99349ff0a86-Abstract-Conference.html}
}

@inproceedings{horwitz2023btf,
  author      = {Horwitz, Eliahu and Hoshen, Yedid},
  title       = {Back to the Feature: Classical {3D} Features Are (Almost) All You Need for {3D} Anomaly Detection},
  booktitle   = {Proceedings of the IEEE/CVF Conference on Computer Vision and Pattern Recognition (CVPR) Workshops},
  year        = {2023},
  pages       = {2968--2977},
  doi         = {10.1109/CVPRW59228.2023.00298},
  url         = {https://openaccess.thecvf.com/content/CVPR2023W/VAND/html/Horwitz_Back_to_the_Feature_Classical_3D_Features_Are_Almost_All_CVPRW_2023_paper.html}
}

@inproceedings{yu2022pointbert,
  author      = {Yu, Xumin and Tang, Lulu and Rao, Yongming and Huang, Tiejun and Zhou, Jie and Lu, Jiwen},
  title       = {{Point-BERT}: Pre-Training {3D} Point Cloud Transformers with Masked Point Modeling},
  booktitle   = {Proceedings of the IEEE/CVF Conference on Computer Vision and Pattern Recognition (CVPR)},
  year        = {2022},
  pages       = {19313--19322},
  url         = {https://openaccess.thecvf.com/content/CVPR2022/html/Yu_Point-BERT_Pre-Training_3D_Point_Cloud_Transformers_With_Masked_Point_Modeling_CVPR_2022_paper.html}
}

@inproceedings{rusu2009fpfh,
  author      = {Rusu, Radu Bogdan and Blodow, Nico and Beetz, Michael},
  title       = {Fast Point Feature Histograms ({FPFH}) for {3D} Registration},
  booktitle   = {Proceedings of the IEEE International Conference on Robotics and Automation (ICRA)},
  year        = {2009},
  pages       = {3212--3217},
  doi         = {10.1109/ROBOT.2009.5152473},
  url         = {https://doi.org/10.1109/ROBOT.2009.5152473}
}

@misc{oord2018representation,
  author       = {van den Oord, Aaron and Li, Yazhe and Vinyals, Oriol},
  title        = {Representation Learning with Contrastive Predictive Coding},
  year         = {2018},
  howpublished = {arXiv preprint arXiv:1807.03748},
  doi          = {10.48550/arXiv.1807.03748},
  url          = {https://arxiv.org/abs/1807.03748}
}

@inproceedings{chen2020simple,
  author      = {Chen, Ting and Kornblith, Simon and Norouzi, Mohammad and Hinton, Geoffrey},
  title       = {A Simple Framework for Contrastive Learning of Visual Representations},
  booktitle   = {Proceedings of the 37th International Conference on Machine Learning (ICML)},
  series      = {Proceedings of Machine Learning Research},
  volume      = {119},
  publisher   = {PMLR},
  year        = {2020},
  pages       = {1597--1607},
  url         = {https://proceedings.mlr.press/v119/chen20j.html}
}

@inproceedings{reiss2023mean,
  author      = {Reiss, Tal and Hoshen, Yedid},
  title       = {Mean-Shifted Contrastive Loss for Anomaly Detection},
  booktitle   = {Proceedings of the AAAI Conference on Artificial Intelligence (AAAI)},
  volume      = {37},
  number      = {2},
  year        = {2023},
  pages       = {2155--2162},
  doi         = {10.1609/aaai.v37i2.25309},
  url         = {https://ojs.aaai.org/index.php/AAAI/article/view/25309}
}

@inproceedings{zhou2024r3dad,
  author      = {Zhou, Zheyuan and Wang, Le and Fang, Naiyu and Wang, Zili and Qiu, Lemiao and Zhang, Shuyou},
  title       = {{R3D-AD}: Reconstruction via Diffusion for {3D} Anomaly Detection},
  booktitle   = {Computer Vision -- ECCV 2024},
  series      = {Lecture Notes in Computer Science},
  volume      = {15094},
  publisher   = {Springer},
  year        = {2025},
  pages       = {91--107},
  doi         = {10.1007/978-3-031-72764-1_6},
  url         = {https://doi.org/10.1007/978-3-031-72764-1_6}
}

@inproceedings{Kruse_2024_CVPR,
  author      = {Kruse, Mathis and Rudolph, Marco and Woiwode, Dominik and Rosenhahn, Bodo},
  title       = {{SplatPose} \& Detect: Pose-Agnostic {3D} Anomaly Detection},
  booktitle   = {Proceedings of the IEEE/CVF Conference on Computer Vision and Pattern Recognition (CVPR) Workshops},
  year        = {2024},
  pages       = {3950--3960},
  doi         = {10.1109/CVPRW63382.2024.00399},
  url         = {https://openaccess.thecvf.com/content/CVPR2024W/VAND/html/Kruse_SplatPose__Detect_Pose-Agnostic_3D_Anomaly_Detection_CVPRW_2024_paper.html}
}

@article{li_gpad,
  author      = {Li, Min and He, Jinghui and Li, Gang and Li, Jiachen and Wan, Jin and Han, Delong},
  title       = {Multimodal Industrial Anomaly Detection via Geometric Prior},
  journal     = {IEEE Transactions on Circuits and Systems for Video Technology (TCSVT)},
  volume      = {36},
  number      = {3},
  year        = {2026},
  pages       = {2854--2866},
  doi         = {10.1109/TCSVT.2025.3613708},
  url         = {https://doi.org/10.1109/TCSVT.2025.3613708}
}

@article{asad2025_2m3df,
  author      = {Asad, Mujtaba and Azeem, Waqar and Jiang, He and Tayyab Mustafa, Hafiz and Yang, Jie and Liu, Wei},
  title       = {{2M3DF}: Advancing {3D} Industrial Defect Detection With Multi-Perspective Multimodal Fusion Network},
  journal     = {IEEE Transactions on Circuits and Systems for Video Technology (TCSVT)},
  volume      = {35},
  number      = {7},
  year        = {2025},
  pages       = {6803--6815},
  doi         = {10.1109/TCSVT.2025.3536475},
  url         = {https://doi.org/10.1109/TCSVT.2025.3536475}
}

@article{liu2024_pumask,
  author      = {Liu, Hao and Yuan, Hui and Hamzaoui, Raouf and Liu, Qi and Li, Shuai},
  title       = {{PU-Mask}: {3D} Point Cloud Upsampling via an Implicit Virtual Mask},
  journal     = {IEEE Transactions on Circuits and Systems for Video Technology (TCSVT)},
  volume      = {34},
  number      = {7},
  year        = {2024},
  pages       = {6489--6502},
  doi         = {10.1109/TCSVT.2024.3370001},
  url         = {https://doi.org/10.1109/TCSVT.2024.3370001}
}

@article{liu2025_pugsm,
  author      = {Liu, Hao and Yuan, Hui and Hamzaoui, Raouf and Yan, Weiqing},
  title       = {{PU-GSM}: A Latent Geometry-Guided Self-Similarity Model for Point Cloud Upsampling},
  journal     = {IEEE Transactions on Circuits and Systems for Video Technology (TCSVT)},
  volume      = {35},
  number      = {11},
  year        = {2025},
  pages       = {11514--11526},
  doi         = {10.1109/TCSVT.2025.3612698},
  url         = {https://doi.org/10.1109/TCSVT.2025.3612698}
}

@article{ding2021_gcpcu,
  author      = {Ding, Dandan and Qiu, Chi and Liu, Fuchang and Pan, Zhigeng},
  title       = {Point Cloud Upsampling via Perturbation Learning},
  journal     = {IEEE Transactions on Circuits and Systems for Video Technology (TCSVT)},
  volume      = {31},
  number      = {12},
  year        = {2021},
  pages       = {4661--4672},
  doi         = {10.1109/TCSVT.2021.3099106},
  url         = {https://doi.org/10.1109/TCSVT.2021.3099106}
}

\clearpage


\section*{Supplementary material}

# Supplementary Material for "$\mathcal{PC}^2$-AD: Point Cloud Upsampling to Safeguard 3D Anomaly Detection with Resolution-constrained Edge Devices"

Yutong Gu, Yingxi Xie, Kejin Huang, Jian Ning, Hanzhe Liang, Linlin Shen, and Jinbao Wang

## S1. Upsampling Quality Evaluation

To further evaluate the geometric quality of the resolution-compensated point clouds, we compare $\mathcal{PC}^2$-AD with three upsampling baselines, Midpoint, MLS, and EAR, on Anomaly-ShapeNet, Anomaly-ShapeNet-new, and Real3D-AD. Upsampling quality is evaluated using Chamfer Distance (CD) [1] and Hausdorff Distance (HD) [2].

### A. Evaluation Protocol

For each test sample, the original high-resolution point cloud before downsampling serves as the geometric reference. For abnormal samples, the reference therefore retains the original anomalous geometry. All methods receive the same low-resolution input and produce the same number of output points for each sample.

Before evaluation, both the output and reference point clouds are normalized using the centroid and maximum radius of the reference cloud. CD is computed as the sum of the mean squared nearest-neighbor distances in the two directions, whereas HD is computed as the maximum nearest-neighbor Euclidean distance over both directions. Lower values indicate better geometric consistency. Both metrics are computed on the complete output point clouds, including the retained input points and newly generated points.

Results are reported separately for normal, abnormal, and all test samples. The normal/abnormal sample counts are 600/712 for Anomaly-ShapeNet, 180/231 for Anomaly-ShapeNet-new, and 604/602 for Real3D-AD. Within each group, sample-level metrics are first averaged within each category, and the reported Average is then computed as the unweighted mean across categories. The all-sample results are computed using all test samples within each category.

### B. Baselines and Implementation Details

**Midpoint.** This baseline generates candidate points at the midpoints between input points and their nearest neighbors. The initial neighborhood size is eight and is increased to 16, 32, or 64 when additional candidates are required to reach the target output size.

**MLS.** We use Moving Least Squares (MLS) [3] as implemented in PCL 1.12.1, with a second-order polynomial fit and the `SAMPLE_LOCAL_PLANE` upsampling mode. Let $s$ denote the median nearest-neighbor distance of the input cloud in the coordinate system used for generation. The fitting radius and local sampling radius are set to $4s$ and $1.5s$, respectively. The sampling step is initially $0.75s$ and is reduced to $0.5s$ or $0.35s$ when additional candidates are required.

**EAR.** We use the edge-aware point set resampling method [4] as implemented in CGAL 5.4. Input normals are estimated and consistently oriented using 30-point neighborhoods in Open3D 0.17.0. The neighborhood radius is set to $4s$, the sharpness angle to $25^\circ$, and the edge sensitivity to zero.

All three baselines retain the original input points. Duplicate candidates and candidates coinciding with input points are removed, after which the required number of additional points is selected without replacement using a deterministic sample-specific random seed derived from seed 42. For EAR, when the candidate count remains insufficient after deduplication, the requested number of added candidates is increased to 1.1 or 1.25 times the required count. The same parameter settings and candidate-count adjustment rules are applied to all test samples.

### C. Quantitative Results

For all tables below, each entry reports CD/HD, with CD and HD scaled by $10^5$ and $10^2$, respectively. Red marks per-category $\mathcal{PC}^2$-AD values lower than all three baselines, while blue marks the best baseline values when lower than $\mathcal{PC}^2$-AD. Colors are determined separately for each metric using unrounded values; Average values are not color-coded.

*1) Results on Anomaly-ShapeNet:* Table S1 reports the results across 40 categories. $\mathcal{PC}^2$-AD achieves the lowest category-averaged CD and HD for normal, abnormal, and all samples. Over all samples, its CD and HD are $7.2614 \times 10^{-5}$ and 0.02062, respectively, corresponding to reductions of 14.5% and 6.4% relative to Midpoint, the strongest baseline in these two aggregate metrics. Similar improvements are observed for normal and abnormal samples, with CD reduced by 14.7% and 14.4% and HD reduced by 6.0% and 6.6%, respectively.

*2) Results on Anomaly-ShapeNet-new:* Table S2 reports the results across 12 categories. $\mathcal{PC}^2$-AD achieves the lowest category-averaged CD and HD for all three sample groups. Over all samples, its CD and HD are $4.9018 \times 10^{-5}$ and 0.01687, respectively, corresponding to reductions of 18.2% and 10.1% relative to Midpoint. Similar improvements are observed for normal and abnormal samples, with CD reduced by 18.4% and 18.1% and HD reduced by 10.4% and 10.0%, respectively.

*3) Results on Real3D-AD:* Table S3 reports the results on Real3D-AD. $\mathcal{PC}^2$-AD achieves the lowest category-averaged HD for normal, abnormal, and all samples. Over all samples, its HD is 0.01471, which is 7.2% lower than Midpoint and 18.5% lower than EAR. For CD, EAR achieves the lowest all-sample average of $2.0753 \times 10^{-5}$, while $\mathcal{PC}^2$-AD obtains $2.1338 \times 10^{-5}$, which is 2.8% higher than EAR but lower than both Midpoint and MLS. Thus, EAR yields a slightly lower average squared nearest-neighbor discrepancy, whereas $\mathcal{PC}^2$-AD yields a lower average sample-level maximum geometric discrepancy.

## S2. Additional Qualitative Results

Fig. S2 provides additional qualitative examples supplementing Fig. 3 in the main paper. Fig. S3 extends the anomaly localization results in Fig. 4 to Point-BERT, BTF, MC3D-AD, and Reg2Inv.

## S3. Additional Parameter Sensitivity

Fig. S1 complements Fig. 5 in the main paper by reporting sensitivity results for the remaining 17 parameters under the same evaluation protocol. Parameters with the same generic name in different training stages are treated as distinct parameters. The panels are ordered by CNP, MSCF, TDA, GACF, and NPPC's same-anchor replacement, from left to right and from top to bottom.

For CNP, the default **Coordinate Jitter** of $0.002/0.010$ achieves the best overall detection performance among the tested settings. A **Learning Rate** of $5 \times 10^{-4}$ slightly improves the aggregate detection metrics over the default $1 \times 10^{-3}$, while the default maintains higher FPS. The preferred **Scaling Range** is 0.90–1.10. For **Temperature**, 0.50 slightly improves the aggregate detection metrics over the default 0.25, whereas the default provides higher FPS. For **Weight Decay**, $1 \times 10^{-4}$ gives a slightly higher Avg. AUROC, while the default $5 \times 10^{-5}$ yields a higher Avg. AUPRC.

For MSCF, the preferred **Maximum Point-Dropout Ratio** and **Coordinate Jitter** are 0.10 and $0.002/0.010$, respectively. The default **Learning Rate** of $1 \times 10^{-3}$ and **Weight Decay** of $5 \times 10^{-5}$ achieve the best overall detection performance among the tested settings. These results indicate that the adopted MSCF configuration remains favorable when its additional augmentation and optimization parameters are varied.

For TDA, the default **Learning Rate** of $1 \times 10^{-5}$ achieves the best overall detection performance among the tested settings. A **Weight Decay** of $1 \times 10^{-4}$ slightly improves the aggregate detection metrics, while the default value of 0 provides comparable performance with higher FPS.

For GACF, the preferred **Strict Upper Distance Bound** $\beta_{\mathrm{str}}$, **Relaxed Upper Distance Bound** $\beta_{\mathrm{rel}}$, and **Regional Scale** $\gamma$ are 2.5, 3.0, and 4.0, respectively. These settings provide the best overall detection performance among the tested values, supporting the adopted upper-distance constraints and regional allocation scale.

For NPPC's same-anchor replacement, the preferred **Candidate Rank-Gap Ratio** $\lambda_{\mathrm{gap}}$, **Strong Rank-Improvement Threshold** $\kappa_{\mathrm{s}}$, and **Strong Rank-Improvement Ratio** $\lambda_{\mathrm{imp}}$ are 0.50, 5000, and 0.50, respectively.

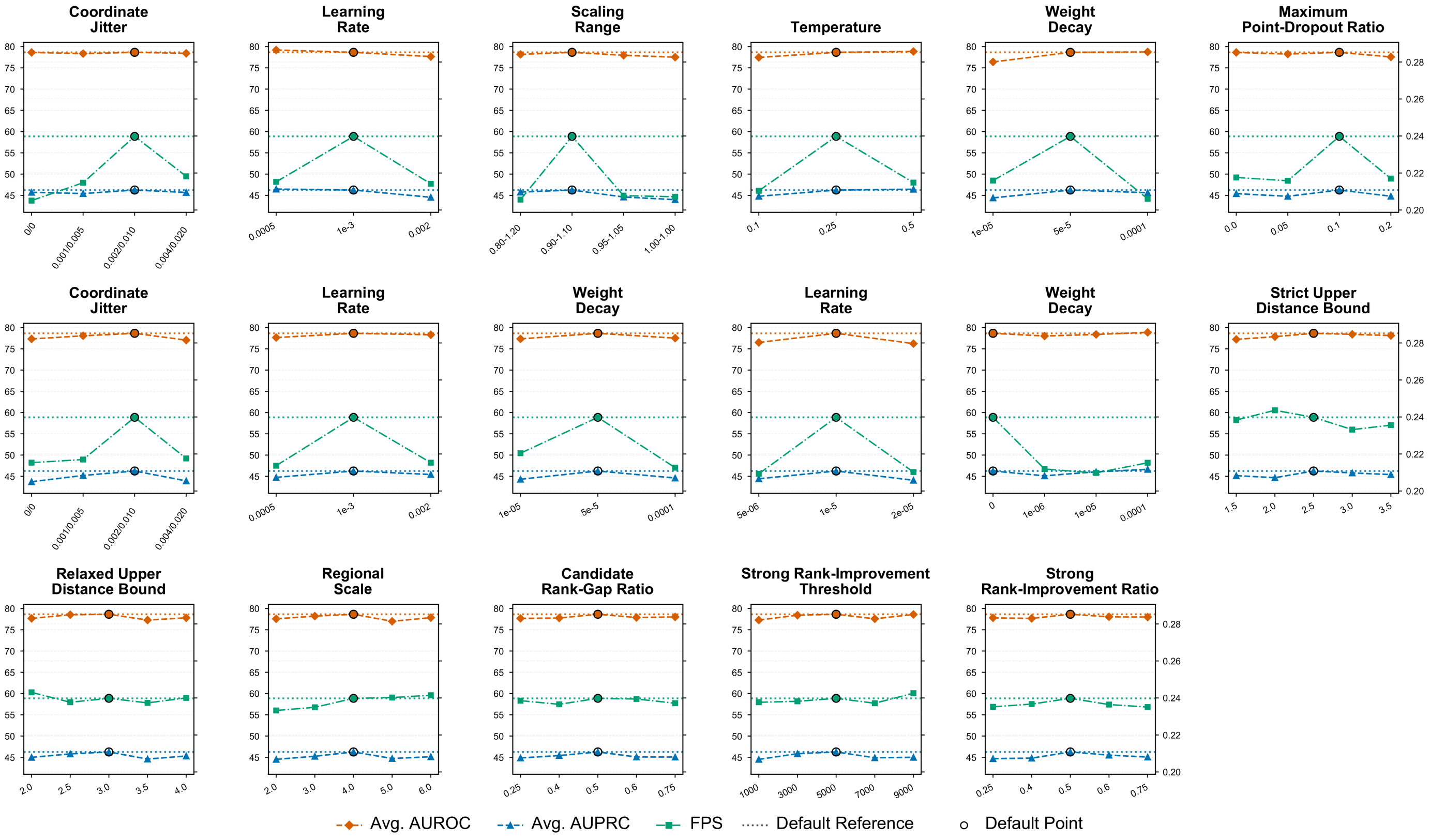


Fig. S1. Additional parameter sensitivity results of $\mathcal{PC}^2$-AD, supplementing Fig. 5 in the main paper, covering the remaining 17 parameters.

TABLE S1

PER-CATEGORY UPSAMPLING QUALITY ON ANOMALY-SHAPENET: (A) NORMAL, (B) ABNORMAL, AND (C) ALL.

**(a) Normal**: CD ($\times 10^5$) / HD ($\times 10^2$) ↓

| Method | ashtray0 | bag0 | bottle0 | bottle1 | bottle3 | bowl0 | bowl1 | bowl2 | bowl3 | bowl4 | bowl5 | bucket0 | bucket1 | cap0 |
|---|---|---|---|---|---|---|---|---|---|---|---|---|---|---|
| Midpoint | 19.90/2.89 | 10.00/2.27 | 1.65/0.87 | 2.50/1.33 | 7.78/1.97 | 9.90/2.22 | 16.34/3.15 | 6.30/2.03 | 9.10/2.11 | 10.95/3.26 | 15.51/2.88 | 9.12/2.04 | 13.69/3.01 | 8.20/2.05 |
| MLS | 42.62/4.07 | 10.46/2.28 | 1.64/0.87 | 2.57/2.04 | 7.92/1.97 | 10.20/2.21 | 61.88/4.31 | 32.33/3.77 | 83.07/4.43 | 67.91/4.35 | 42.76/4.37 | 31.73/4.25 | 61.38/3.99 | 85.17/4.01 |
| EAR | 38.15/4.05 | 9.39/1.87 | 1.47/0.75 | 2.27/1.29 | 7.32/1.93 | 8.89/1.65 | 41.74/4.35 | 17.28/3.10 | 27.53/4.49 | 44.29/3.54 | 29.15/5.04 | 13.81/4.76 | 47.42/5.25 | 30.05/4.11 |
| $\mathcal{PC}^2$-AD | 17.81/2.75 | 8.85/2.13 | 1.40/0.86 | 2.12/1.25 | 6.85/1.94 | 8.50/2.20 | 13.44/2.82 | 5.15/1.78 | 7.78/2.08 | 9.14/2.91 | 13.52/2.62 | 7.70/2.03 | 10.44/2.89 | 7.02/2.01 |

| Method | cap3 | cap4 | cap5 | cup0 | cup1 | eraser0 | headset0 | headset1 | helmet0 | helmet1 | helmet2 | helmet3 | jar0 | microphone0 |
|---|---|---|---|---|---|---|---|---|---|---|---|---|---|---|
| Midpoint | 3.76/1.54 | 4.43/1.51 | 3.95/1.20 | 7.33/1.91 | 15.89/2.53 | 10.10/2.25 | 2.41/1.08 | 9.34/2.93 | 3.45/1.57 | 15.29/5.51 | 9.47/2.89 | 10.51/2.44 | 10.88/2.27 | 2.91/1.25 |
| MLS | 12.69/2.00 | 14.03/2.31 | 9.93/1.97 | 25.43/3.70 | 37.41/3.67 | 11.16/2.26 | 6.24/1.55 | 14.09/3.60 | 14.26/3.91 | 36.02/7.19 | 53.21/3.68 | 15.64/4.19 | 11.07/2.27 | 3.00/1.25 |
| EAR | 9.92/1.80 | 10.83/1.92 | 8.79/1.21 | 9.60/4.43 | 32.27/2.71 | 8.58/1.61 | 6.58/4.77 | 11.06/3.13 | 10.45/5.53 | 28.46/6.72 | 38.17/3.02 | 11.82/4.55 | 9.70/1.87 | 2.68/0.92 |
| $\mathcal{PC}^2$-AD | 3.18/1.44 | 3.76/1.37 | 3.36/1.18 | 6.22/1.89 | 14.04/2.42 | 8.78/2.10 | 2.16/1.00 | 8.09/2.52 | 2.88/1.47 | 13.27/5.52 | 7.37/2.67 | 8.90/2.33 | 9.49/2.21 | 2.56/1.24 |

| Method | shelf0 | tap0 | tap1 | vase0 | vase1 | vase2 | vase3 | vase4 | vase5 | vase7 | vase8 | vase9 | Average |
|---|---|---|---|---|---|---|---|---|---|---|---|---|---|
| Midpoint | 6.82/1.64 | 2.99/1.23 | 2.78/1.20 | 2.79/1.12 | 6.47/2.52 | 11.25/2.35 | 6.45/2.56 | 4.42/1.31 | 17.90/2.69 | 9.12/3.04 | 9.29/2.94 | 5.13/1.59 | 8.40/2.18 |
| MLS | 17.10/3.08 | 7.20/1.69 | 7.10/1.91 | 8.46/1.56 | 57.91/3.12 | 11.53/2.35 | 47.26/2.76 | 10.73/1.85 | 41.90/3.50 | 53.39/4.54 | 62.71/3.51 | 13.13/1.76 | 28.61/3.05 |
| EAR | 10.33/3.65 | 5.55/1.30 | 5.09/1.84 | 6.94/1.39 | 28.30/3.44 | 10.25/1.85 | 28.03/3.44 | 9.45/1.40 | 35.85/2.68 | 20.89/5.10 | 43.25/3.03 | 10.60/1.74 | 18.30/3.03 |
| $\mathcal{PC}^2$-AD | 5.72/1.63 | 2.49/1.17 | 2.36/1.11 | 2.33/1.07 | 5.58/1.97 | 9.82/2.34 | 5.33/2.48 | 3.74/1.24 | 15.84/2.58 | 7.79/2.29 | 7.47/2.84 | 4.38/1.51 | 7.17/2.05 |

**(b) Abnormal**: CD ($\times 10^5$) / HD ($\times 10^2$) ↓

| Method | ashtray0 | bag0 | bottle0 | bottle1 | bottle3 | bowl0 | bowl1 | bowl2 | bowl3 | bowl4 | bowl5 | bucket0 | bucket1 | cap0 |
|---|---|---|---|---|---|---|---|---|---|---|---|---|---|---|
| Midpoint | 18.90/2.76 | 9.96/2.28 | 1.64/0.88 | 3.77/1.42 | 10.76/2.13 | 9.86/2.23 | 16.29/3.23 | 6.20/2.12 | 9.12/2.12 | 11.01/3.28 | 15.41/2.98 | 12.27/2.39 | 13.65/3.03 | 8.27/2.57 |
| MLS | 39.28/3.81 | 10.49/2.29 | 1.63/0.94 | 6.24/2.00 | 14.46/2.34 | 10.36/2.34 | 61.14/4.28 | 31.44/3.15 | 82.77/4.45 | 69.02/4.17 | 42.94/4.40 | 30.80/4.38 | 61.20/4.17 | 84.22/4.12 |
| EAR | 35.23/3.90 | 9.42/1.89 | 1.46/0.83 | 5.23/1.55 | 12.42/2.12 | 8.89/1.93 | 41.57/4.19 | 16.92/3.06 | 27.98/4.53 | 43.59/3.72 | 29.58/5.34 | 18.22/4.77 | 46.88/5.40 | 29.94/4.24 |
| $\mathcal{PC}^2$-AD | 16.78/2.64 | 8.84/2.17 | 1.39/0.87 | 3.26/1.35 | 9.65/2.11 | 8.44/2.21 | 13.57/2.88 | 5.07/1.79 | 7.81/2.09 | 9.15/2.91 | 13.39/2.70 | 10.81/2.27 | 10.50/2.89 | 7.06/2.35 |

| Method | cap3 | cap4 | cap5 | cup0 | cup1 | eraser0 | headset0 | headset1 | helmet0 | helmet1 | helmet2 | helmet3 | jar0 | microphone0 |
|---|---|---|---|---|---|---|---|---|---|---|---|---|---|---|
| Midpoint | 3.86/1.57 | 4.32/1.62 | 3.93/1.15 | 7.37/2.06 | 15.88/2.56 | 10.15/2.29 | 2.42/1.09 | 9.36/2.92 | 3.42/1.56 | 15.40/5.56 | 9.58/2.87 | 10.50/2.37 | 10.79/2.26 | 2.94/1.28 |
| MLS | 13.32/2.88 | 12.87/2.20 | 9.17/1.87 | 27.16/3.82 | 37.42/3.80 | 11.24/2.51 | 6.27/1.57 | 14.12/3.65 | 14.04/4.39 | 67.43/92.43 | 51.97/3.79 | 15.55/4.06 | 10.98/2.34 | 3.01/1.38 |
| EAR | 10.21/2.17 | 10.41/2.19 | 8.39/1.30 | 10.38/4.45 | 32.22/2.89 | 8.69/1.93 | 6.58/4.82 | 11.10/3.31 | 10.70/5.68 | 28.85/6.85 | 38.59/3.36 | 11.82/4.72 | 9.66/1.96 | 2.72/1.12 |
| $\mathcal{PC}^2$-AD | 3.23/1.50 | 3.71/1.35 | 3.38/1.13 | 6.24/1.93 | 14.03/2.39 | 8.85/2.20 | 2.09/1.00 | 8.10/2.45 | 2.86/1.45 | 13.54/5.56 | 7.41/2.64 | 8.91/2.27 | 9.42/2.20 | 2.61/1.24 |

| Method | shelf0 | tap0 | tap1 | vase0 | vase1 | vase2 | vase3 | vase4 | vase5 | vase7 | vase8 | vase9 | Average |
|---|---|---|---|---|---|---|---|---|---|---|---|---|---|
| Midpoint | 6.96/1.76 | 3.01/1.22 | 2.81/1.19 | 2.75/1.10 | 6.42/2.43 | 11.25/2.36 | 6.43/2.52 | 4.41/1.31 | 17.77/2.66 | 9.09/3.17 | 9.33/2.96 | 5.14/1.61 | 8.56/2.22 |
| MLS | 16.35/3.07 | 7.06/1.69 | 6.72/1.73 | 7.91/1.48 | 57.39/3.21 | 11.73/2.37 | 45.57/2.95 | 10.74/2.01 | 41.62/3.48 | 53.29/4.51 | 62.76/3.76 | 13.60/1.86 | 29.38/5.24 |
| EAR | 10.31/3.70 | 5.51/1.81 | 4.96/1.91 | 6.76/1.40 | 28.31/3.37 | 10.27/1.92 | 27.13/3.75 | 9.43/1.46 | 35.75/2.72 | 21.14/5.12 | 43.45/3.65 | 10.76/2.38 | 18.54/3.19 |
| $\mathcal{PC}^2$-AD | 5.85/1.71 | 2.51/1.17 | 2.45/1.11 | 2.30/1.06 | 5.57/1.95 | 9.82/2.29 | 5.31/2.43 | 3.74/1.24 | 15.75/2.56 | 7.77/2.52 | 7.47/2.83 | 4.41/1.57 | 7.33/2.07 |

**(c) All**: CD ($\times 10^5$) / HD ($\times 10^2$) ↓

| Method | ashtray0 | bag0 | bottle0 | bottle1 | bottle3 | bowl0 | bowl1 | bowl2 | bowl3 | bowl4 | bowl5 | bucket0 | bucket1 | cap0 |
|---|---|---|---|---|---|---|---|---|---|---|---|---|---|---|
| Midpoint | 19.42/2.83 | 9.98/2.27 | 1.65/0.87 | 3.21/1.38 | 9.52/2.07 | 9.88/2.22 | 16.31/3.19 | 6.24/2.08 | 9.11/2.11 | 10.98/3.27 | 15.46/2.94 | 10.96/2.25 | 13.67/3.02 | 8.23/2.33 |
| MLS | 41.00/3.94 | 10.47/2.28 | 1.63/0.90 | 4.62/2.01 | 11.73/2.18 | 10.28/2.28 | 61.48/4.29 | 31.85/3.43 | 82.90/4.44 | 68.52/4.25 | 42.86/4.39 | 31.19/4.33 | 61.28/4.10 | 84.65/4.07 |
| EAR | 36.74/3.98 | 9.41/1.88 | 1.47/0.79 | 3.93/1.44 | 10.30/2.04 | 8.89/1.80 | 41.65/4.27 | 17.08/3.08 | 27.78/4.51 | 43.91/3.64 | 29.39/5.21 | 16.38/4.77 | 47.11/5.33 | 29.99/4.18 |
| $\mathcal{PC}^2$-AD | 17.32/2.70 | 8.85/2.15 | 1.40/0.87 | 2.76/1.31 | 8.49/2.04 | 8.47/2.20 | 13.51/2.85 | 5.11/1.79 | 7.79/2.09 | 9.14/2.91 | 13.44/2.66 | 9.52/2.17 | 10.47/2.89 | 7.04/2.20 |

| Method | cap3 | cap4 | cap5 | cup0 | cup1 | eraser0 | headset0 | headset1 | helmet0 | helmet1 | helmet2 | helmet3 | jar0 | microphone0 |
|---|---|---|---|---|---|---|---|---|---|---|---|---|---|---|
| Midpoint | 3.81/1.56 | 4.37/1.57 | 3.94/1.17 | 7.35/1.99 | 15.89/2.55 | 10.12/2.27 | 2.42/1.08 | 9.35/2.92 | 3.43/1.56 | 15.34/5.53 | 9.54/2.88 | 10.50/2.40 | 10.84/2.27 | 2.92/1.26 |
| MLS | 13.04/2.49 | 13.38/2.25 | 9.50/1.91 | 26.27/3.75 | 37.41/3.74 | 11.20/2.38 | 6.26/1.56 | 14.10/3.63 | 14.13/4.20 | 51.18/48.34 | 52.46/3.75 | 15.59/4.12 | 11.02/2.30 | 3.00/1.32 |
| EAR | 10.08/2.01 | 10.60/2.07 | 8.57/1.26 | 9.97/4.44 | 32.24/2.80 | 8.63/1.76 | 6.58/4.79 | 11.08/3.22 | 10.60/5.62 | 28.65/6.79 | 38.42/3.22 | 11.82/4.65 | 9.68/1.91 | 2.70/1.02 |
| $\mathcal{PC}^2$-AD | 3.21/1.47 | 3.73/1.36 | 3.37/1.15 | 6.23/1.91 | 14.04/2.41 | 8.82/2.15 | 2.13/1.00 | 8.09/2.49 | 2.87/1.46 | 13.40/5.54 | 7.40/2.65 | 8.91/2.29 | 9.46/2.20 | 2.59/1.24 |

| Method | shelf0 | tap0 | tap1 | vase0 | vase1 | vase2 | vase3 | vase4 | vase5 | vase7 | vase8 | vase9 | Average |
|---|---|---|---|---|---|---|---|---|---|---|---|---|---|
| Midpoint | 6.91/1.71 | 3.01/1.23 | 2.80/1.19 | 2.77/1.11 | 6.45/2.48 | 11.25/2.36 | 6.44/2.53 | 4.42/1.31 | 17.84/2.68 | 9.10/3.10 | 9.31/2.95 | 5.14/1.60 | 8.50/2.20 |
| MLS | 16.65/3.08 | 7.11/1.69 | 6.89/1.81 | 8.17/1.52 | 57.66/3.31 | 11.63/2.36 | 46.25/2.87 | 10.74/1.95 | 41.76/3.49 | 53.34/4.53 | 62.74/3.66 | 13.41/1.82 | 28.98/4.11 |
| EAR | 10.32/3.68 | 5.52/1.60 | 5.02/1.88 | 6.85/1.39 | 28.30/3.41 | 10.26/1.88 | 27.50/3.62 | 9.44/1.43 | 35.80/2.70 | 21.01/5.11 | 43.37/3.40 | 10.70/2.12 | 18.44/3.12 |
| $\mathcal{PC}^2$-AD | 5.80/1.68 | 2.50/1.17 | 2.41/1.11 | 2.32/1.06 | 5.58/1.96 | 9.82/2.32 | 5.32/2.45 | 3.74/1.24 | 15.80/2.57 | 7.78/2.40 | 7.47/2.83 | 4.39/1.55 | 7.26/2.06 |

TABLE S2

PER-CATEGORY UPSAMPLING QUALITY ON ANOMALY-SHAPENET-NEW: (A) NORMAL, (B) ABNORMAL, AND (C) ALL.

**(a) Normal**: CD ($\times 10^5$) / HD ($\times 10^2$) ↓

| Method | cabinet0 | cap1 | cap2 | chair0 | cup2 | desk0 | knife0 | knife1 | microphone1 | screen0 | vase10 | vase6 | Average |
|---|---|---|---|---|---|---|---|---|---|---|---|---|---|
| Midpoint | 9.18/2.99 | 5.55/1.93 | 3.45/1.27 | 12.09/2.39 | 10.30/2.44 | 2.24/1.02 | 0.89/0.74 | 1.05/0.93 | 2.31/1.08 | 2.71/1.15 | 11.38/3.11 | 9.96/3.28 | 5.92/1.86 |
| MLS | 23.68/3.26 | 10.84/2.96 | 10.68/1.87 | 13.32/2.45 | 61.46/4.74 | 2.32/1.02 | 1.10/1.22 | 2.42/1.29 | 2.44/1.08 | 2.56/1.15 | 22.76/4.33 | 86.01/4.65 | 19.97/2.50 |
| EAR | 10.55/3.21 | 7.06/2.15 | 8.55/1.68 | 8.59/3.25 | 24.57/5.04 | 1.42/1.06 | 0.63/1.09 | 1.34/1.42 | 2.17/0.88 | 1.63/1.09 | 13.12/4.41 | 36.08/4.99 | 9.64/2.52 |
| $\mathcal{PC}^2$-AD | 6.93/2.18 | 4.88/1.79 | 2.89/1.20 | 9.04/2.33 | 8.74/2.22 | 1.70/1.02 | 0.69/0.71 | 0.83/0.86 | 1.99/1.05 | 2.06/1.15 | 9.74/2.87 | 8.52/2.61 | 4.84/1.67 |

**(b) Abnormal**: CD ($\times 10^5$) / HD ($\times 10^2$) ↓

| Method | cabinet0 | cap1 | cap2 | chair0 | cup2 | desk0 | knife0 | knife1 | microphone1 | screen0 | vase10 | vase6 | Average |
|---|---|---|---|---|---|---|---|---|---|---|---|---|---|
| Midpoint | 9.17/2.97 | 5.13/1.93 | 3.44/1.22 | 12.11/2.43 | 10.23/2.59 | 2.24/1.07 | 0.89/0.76 | 1.05/0.95 | 2.31/1.08 | 2.71/1.17 | 12.79/3.26 | 10.33/3.28 | 6.03/1.89 |
| MLS | 23.38/3.27 | 9.69/2.71 | 9.92/1.82 | 13.10/2.49 | 60.95/4.67 | 2.32/1.14 | 1.11/1.18 | 2.42/1.30 | 2.43/1.08 | 2.66/1.27 | 26.69/4.54 | 84.48/4.71 | 19.93/2.51 |
| EAR | 10.74/3.45 | 6.57/2.29 | 8.35/1.92 | 8.64/3.40 | 24.47/5.03 | 1.43/1.36 | 0.65/1.26 | 1.37/1.49 | 2.17/0.95 | 1.63/1.57 | 15.89/4.87 | 36.68/4.99 | 9.88/2.72 |
| $\mathcal{PC}^2$-AD | 6.96/2.16 | 4.51/1.86 | 2.90/1.18 | 9.05/2.40 | 8.64/2.25 | 1.71/1.04 | 0.70/0.71 | 0.83/0.89 | 1.99/1.07 | 2.06/1.17 | 11.10/2.97 | 8.86/2.75 | 4.94/1.70 |

**(c) All**: CD ($\times 10^5$) / HD ($\times 10^2$) ↓

| Method | cabinet0 | cap1 | cap2 | chair0 | cup2 | desk0 | knife0 | knife1 | microphone1 | screen0 | vase10 | vase6 | Average |
|---|---|---|---|---|---|---|---|---|---|---|---|---|---|
| Midpoint | 9.17/2.98 | 5.31/1.93 | 3.44/1.24 | 12.10/2.41 | 10.27/2.51 | 2.24/1.05 | 0.89/0.75 | 1.05/0.94 | 2.31/1.08 | 2.71/1.16 | 12.23/3.20 | 10.18/3.28 | 5.99/1.88 |
| MLS | 23.51/3.26 | 10.19/2.82 | 10.26/1.84 | 13.20/2.47 | 61.22/4.71 | 2.32/1.09 | 1.10/1.20 | 2.42/1.30 | 2.43/1.08 | 2.61/1.21 | 25.14/4.46 | 85.10/4.68 | 19.96/2.51 |
| EAR | 10.66/3.34 | 6.79/2.23 | 8.44/1.82 | 8.61/3.33 | 24.52/5.04 | 1.43/1.23 | 0.64/1.19 | 1.36/1.46 | 2.17/0.92 | 1.63/1.35 | 14.80/4.69 | 36.44/4.99 | 9.79/2.63 |
| $\mathcal{PC}^2$-AD | 6.95/2.17 | 4.67/1.83 | 2.90/1.19 | 9.05/2.37 | 8.69/2.24 | 1.71/1.03 | 0.69/0.71 | 0.83/0.88 | 1.99/1.06 | 2.06/1.16 | 10.56/2.93 | 8.72/2.70 | 4.90/1.69 |

TABLE S3

PER-CATEGORY UPSAMPLING QUALITY ON REAL3D-AD: (A) NORMAL, (B) ABNORMAL, AND (C) ALL.

**(a) Normal**: CD ($\times 10^5$) / HD ($\times 10^2$) ↓

| Method | Airplane | Car | Candybar | Chicken | Diamond | Duck | Fish | Gemstone | Seahorse | Shell | Starfish | Toffees | Average |
|---|---|---|---|---|---|---|---|---|---|---|---|---|---|
| Midpoint | 0.62/1.22 | 0.79/0.94 | 0.78/1.13 | 0.72/1.62 | 1.87/1.55 | 0.60/1.34 | 0.79/1.04 | 1.47/4.86 | 0.45/0.85 | 1.07/1.36 | 0.67/1.93 | 0.51/0.65 | 0.86/1.54 |
| MLS | 0.61/1.77 | 0.77/1.09 | 0.79/1.00 | 1.21/34.58 | 1.87/1.65 | 0.59/1.13 | 0.80/1.02 | 1.29/2.26 | 0.46/0.95 | 1.06/1.25 | 0.67/0.95 | 0.51/0.77 | 0.89/4.04 |
| EAR | 0.50/2.19 | 0.60/1.34 | 0.60/1.77 | 0.61/1.80 | 1.33/1.95 | 0.50/2.08 | 0.56/1.79 | 1.09/1.87 | 0.39/1.63 | 0.90/2.34 | 0.55/1.80 | 0.35/0.77 | 0.67/1.78 |
| $\mathcal{PC}^2$-AD | 0.52/1.21 | 0.67/0.90 | 0.65/1.05 | 0.61/1.59 | 1.53/1.49 | 0.51/1.30 | 0.64/1.00 | 1.27/3.97 | 0.38/0.81 | 0.92/1.34 | 0.57/1.69 | 0.43/0.61 | 0.72/1.41 |

**(b) Abnormal**: CD ($\times 10^5$) / HD ($\times 10^2$) ↓

| Method | Airplane | Car | Candybar | Chicken | Diamond | Duck | Fish | Gemstone | Seahorse | Shell | Starfish | Toffees | Average |
|---|---|---|---|---|---|---|---|---|---|---|---|---|---|
| Midpoint | 0.40/1.25 | 0.78/0.90 | 0.77/1.30 | 33.96/3.30 | 1.87/1.49 | 0.59/1.13 | 0.80/1.01 | 1.64/4.19 | 0.45/0.86 | 1.05/2.07 | 0.70/1.38 | 0.50/0.65 | 3.63/1.63 |
| MLS | 0.39/12.21 | 0.76/1.10 | 0.77/0.95 | 33.40/27.69 | 1.86/1.48 | 0.58/1.01 | 0.80/1.07 | 1.48/1.47 | 0.46/1.15 | 1.05/1.27 | 0.71/1.18 | 0.50/0.81 | 3.56/4.28 |
| EAR | 0.34/1.60 | 0.59/1.19 | 0.59/1.72 | 33.89/2.94 | 1.35/1.96 | 0.47/2.08 | 0.57/1.83 | 1.16/1.83 | 0.39/1.59 | 0.87/2.45 | 0.63/2.02 | 0.35/0.75 | 3.43/1.83 |
| $\mathcal{PC}^2$-AD | 0.34/1.25 | 0.66/0.85 | 0.64/1.24 | 33.87/3.27 | 1.53/1.43 | 0.50/1.09 | 0.65/0.98 | 1.38/3.45 | 0.38/0.83 | 0.90/2.04 | 0.62/1.30 | 0.42/0.62 | 3.49/1.53 |

**(c) All**: CD ($\times 10^5$) / HD ($\times 10^2$) ↓

| Method | Airplane | Car | Candybar | Chicken | Diamond | Duck | Fish | Gemstone | Seahorse | Shell | Starfish | Toffees | Average |
|---|---|---|---|---|---|---|---|---|---|---|---|---|---|
| Midpoint | 0.51/1.24 | 0.79/0.92 | 0.78/1.22 | 17.66/2.48 | 1.87/1.52 | 0.59/1.24 | 0.80/1.03 | 1.56/4.53 | 0.45/0.86 | 1.06/1.70 | 0.69/1.66 | 0.51/0.65 | 2.27/1.59 |
| MLS | 0.50/6.99 | 0.76/1.09 | 0.78/0.98 | 17.61/31.07 | 1.86/1.57 | 0.58/1.07 | 0.80/1.05 | 1.39/1.87 | 0.46/1.05 | 1.06/1.26 | 0.69/1.07 | 0.50/0.79 | 2.25/4.15 |
| EAR | 0.42/1.90 | 0.60/1.26 | 0.59/1.75 | 17.56/2.38 | 1.34/1.95 | 0.48/2.08 | 0.56/1.81 | 1.12/1.85 | 0.39/1.61 | 0.89/2.39 | 0.59/1.91 | 0.35/0.76 | 2.08/1.80 |
| $\mathcal{PC}^2$-AD | 0.43/1.23 | 0.67/0.88 | 0.64/1.15 | 17.55/2.45 | 1.53/1.46 | 0.50/1.20 | 0.65/0.99 | 1.33/3.71 | 0.38/0.82 | 0.91/1.67 | 0.59/1.49 | 0.42/0.61 | 2.13/1.47 |

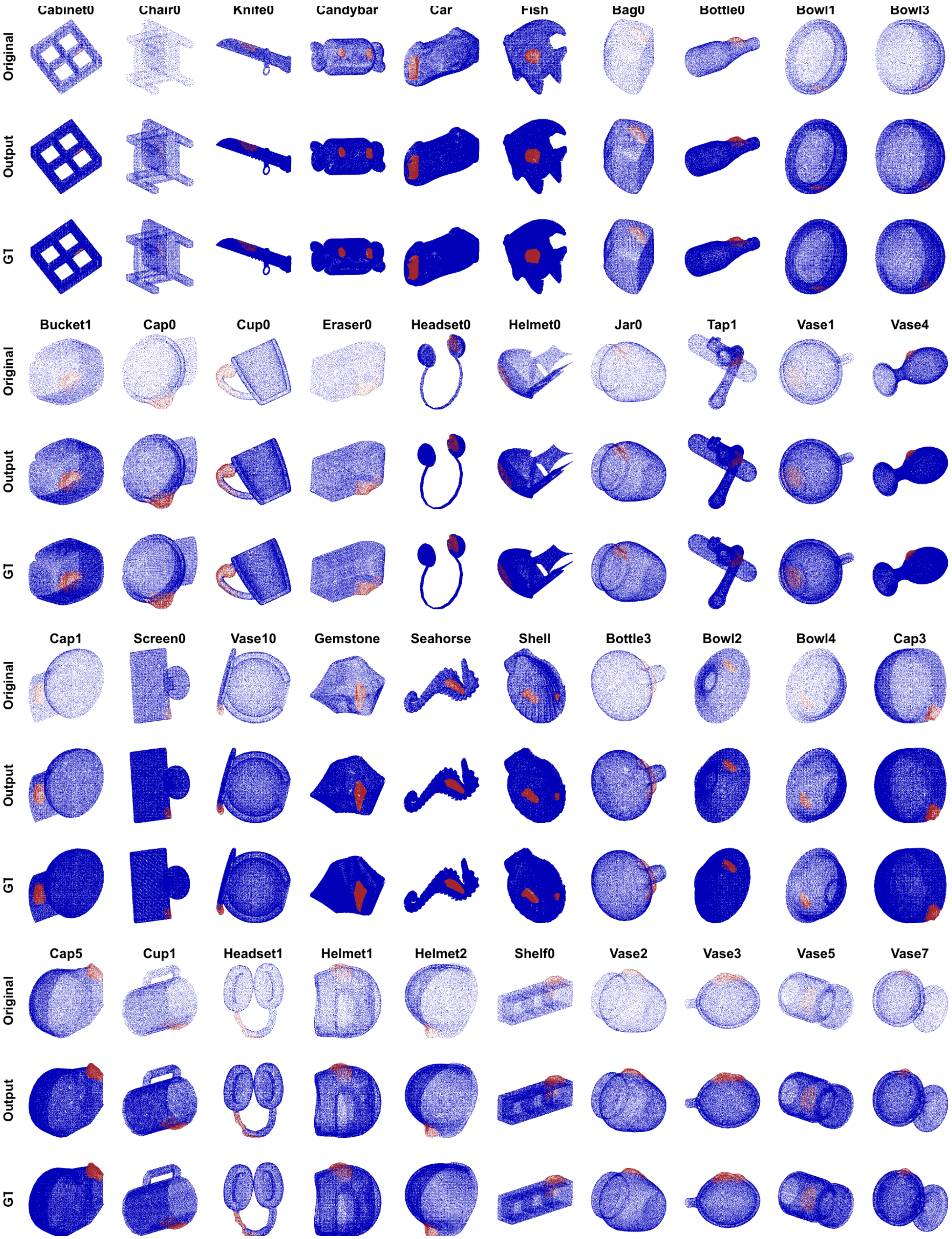


Fig. S2. Additional qualitative results of $\mathcal{PC}^2$-AD, supplementing Fig. 3 in the main paper.

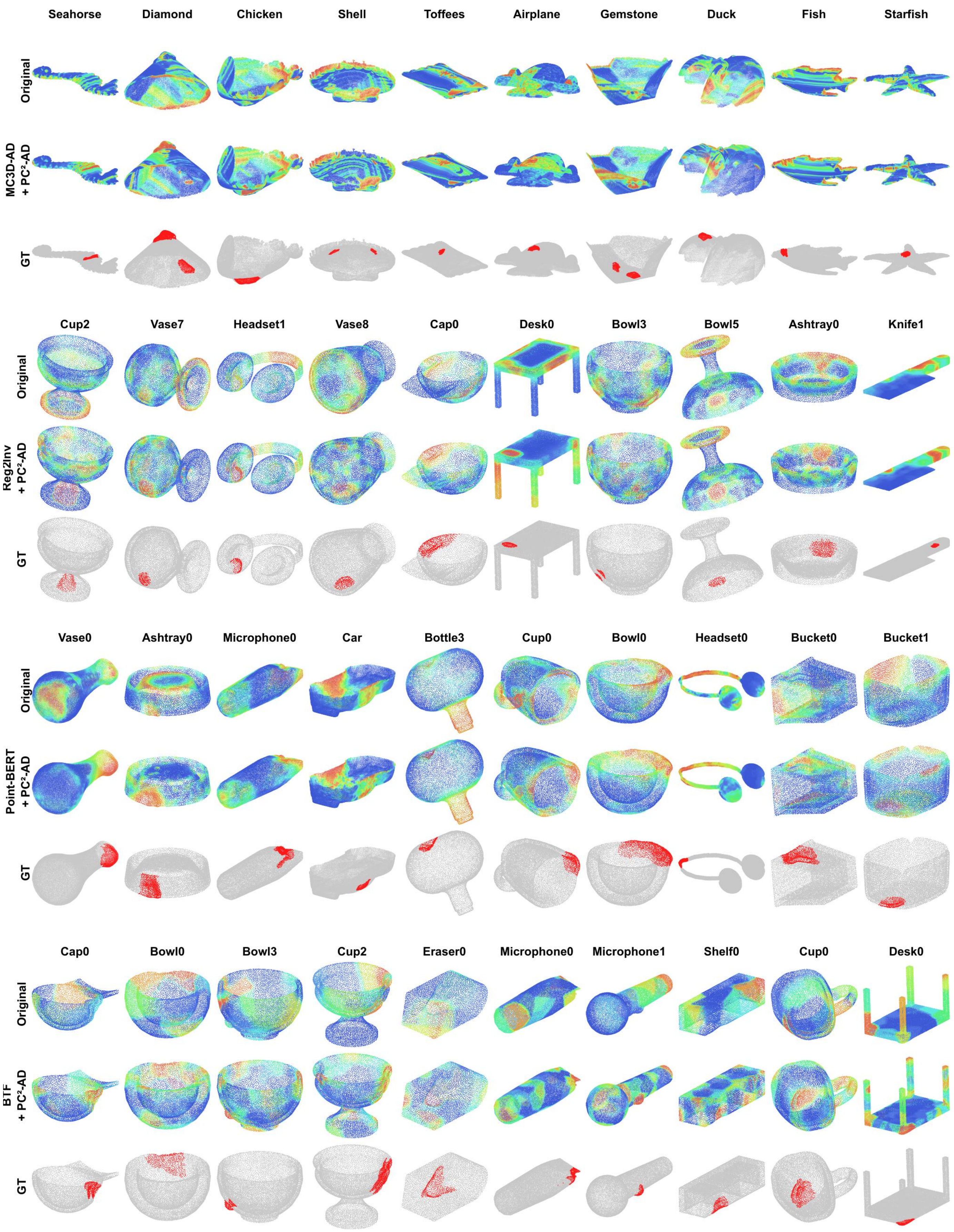


Fig. S3. Additional anomaly localization results, supplementing Fig. 4 in the main paper.

# References


[1] H. Fan, H. Su, and L. J. Guibas, “A point set generation network for 3D object reconstruction from a single image,” in *Proceedings of the IEEE Conference on Computer Vision and Pattern Recognition (CVPR)*, 2017, pp. 605–613. [Online]. Available: https://openaccess.thecvf.com/content_cvpr_2017/html/Fan_A_Point_Set_CVPR_2017_paper.html

[2] D. P. Huttenlocher, G. A. Klanderman, and W. J. Rucklidge, “Comparing images using the Hausdorff distance,” *IEEE Transactions on Pattern Analysis and Machine Intelligence (TPAMI)*, vol. 15, no. 9, pp. 850–863, 1993. [Online]. Available: https://doi.org/10.1109/34.232073

[3] M. Alexa, J. Behr, D. Cohen-Or, S. Fleishman, D. Levin, and C. T. Silva, “Computing and rendering point set surfaces,” *IEEE Transactions on Visualization and Computer Graphics (TVCG)*, vol. 9, no. 1, pp. 3–15, 2003. [Online]. Available: https://doi.org/10.1109/TVCG.2003.1175093

[4] H. Huang, S. Wu, M. Gong, D. Cohen-Or, U. Ascher, and H. R. Zhang, “Edge-aware point set resampling,” *ACM Transactions on Graphics (TOG)*, vol. 32, no. 1, pp. 9:1–9:12, 2013. [Online]. Available: https://doi.org/10.1145/2421636.2421645



\end{document}